\documentclass{article} 
\usepackage{iclr2027_conference,times}

\usepackage{amsmath,amsfonts,bm}

\def\eqref#1{equation~\ref{#1}}

\def\1{\bm{1}}

\DeclareMathAlphabet{\mathsfit}{\encodingdefault}{\sfdefault}{m}{sl}
\SetMathAlphabet{\mathsfit}{bold}{\encodingdefault}{\sfdefault}{bx}{n}

\usepackage{hyperref}
\usepackage{url}
\usepackage[utf8]{inputenc} 
\usepackage[T1]{fontenc}    
\usepackage{hyperref}       
\usepackage{url}            
\usepackage{booktabs}       
\usepackage{amsfonts}       
\usepackage{nicefrac}       
\usepackage{microtype}      
\usepackage{xcolor}         
\usepackage{amsmath}
\usepackage{graphicx}
\usepackage{booktabs}
\usepackage{multirow}
\usepackage[table]{xcolor}
\usepackage{caption}
\usepackage{algorithm}
\usepackage{algorithmic}
\usepackage{titletoc}
\usepackage{amssymb}
\providecommand{\hd}[2]{\begin{tabular}[b]{@{}c@{}}#1\\#2\end{tabular}}
\usepackage[framemethod=tikz]{mdframed}
\newmdenv[
  backgroundcolor=black!3,
  linecolor=black!40,
  linewidth=0.6pt,
  roundcorner=2pt,
  innerleftmargin=8pt,
  innerrightmargin=8pt,
  innertopmargin=10pt,
  innerbottommargin=8pt,
  skipabove=6pt,
  skipbelow=6pt,
  frametitlefont=\normalsize,
  frametitlebackgroundcolor=black!8,
  frametitlerule=true,
  frametitleaboveskip=3pt,
  frametitlebelowskip=3pt
]{promptbox}
\newcommand{\red}[1]{\textcolor{red}{#1}}

\title{Beyond Frame Selection: Rethinking \\Long-Video Understanding with MLLMs}

\iclrfinalcopy
\author{Ziling Huang$^{1}$, Yuki M. Asano$^{2}$, Shin'ichi Satoh$^{1}$\\
$^{1}$National Institute of Informatics, Japan \\
$^{2}$University of Technology Nuremberg, Germany \\
\texttt{\{huangziling, satoh\}@nii.ac.jp}, \texttt{yuki.asano@utn.de} \\
}

\begin{document}

\maketitle

\begin{abstract}
Multimodal Large Language Models (MLLMs) have made strong progress in video
understanding, yet long videos remain difficult: the visual token budget grows
with video length, so temporally sparse evidence is easily lost. Existing methods, whether frame selection or agent-based exploration, ultimately rely on a single frame set that must trade off broad temporal coverage against fine-grained, question-relevant evidence: mixing both dilutes decisive evidence, while focusing on either sacrifices the other. We propose \textbf{VideoRouter} (VR), which rethinks
long-video understanding as coordinating complementary evidence views rather than
selecting a single subset of frames. VideoRouter first organizes each video into
a question-agnostic temporal hierarchy that partitions it into coarse-to-fine
temporally coherent segments. Upper-level nodes capture broad storyline context
and event progression, while lower-level nodes preserve fine-grained local
details and evidence-bearing moments. This gives rise to two complementary views:
a global view for coverage-oriented reasoning and a local view for
detail-oriented evidence recovery. We further introduce a verification-guided
router that judges which view is better supported by its own selected evidence
and decides the final answer. Across six backbones, routing improves over both
views in all settings, and the choice of view is shown to be
dataset-dependent , confirming that no single evidence granularity is universally
preferable. 
On VideoMME and LongVideoBench, our method outperforms state-of-the-art frame-selection methods by 2.5 and 1.3 points, respectively, under the LLaVA-Video-7B backbone.
We will release the code.
\end{abstract}

\section{Introduction}


Multimodal Large Language Models (MLLMs)~\citep{qwen2, qwen2.5, llavanextvideo, llavaonevision, longva, longvila, slowfocus} have made substantial progress in video understanding, yet long-video understanding remains challenging because the amount of visual content can far exceed the available input budget. More importantly, question-relevant evidence may appear at very different temporal scales: some questions require broad coverage of event progression, while others depend on brief, fine-grained details. Effective long-video reasoning therefore requires not only deciding which visual evidence to retain, but also representing that evidence at an appropriate temporal granularity.

\begin{figure}[h]
  \centering
  \includegraphics[width=1.0\linewidth]{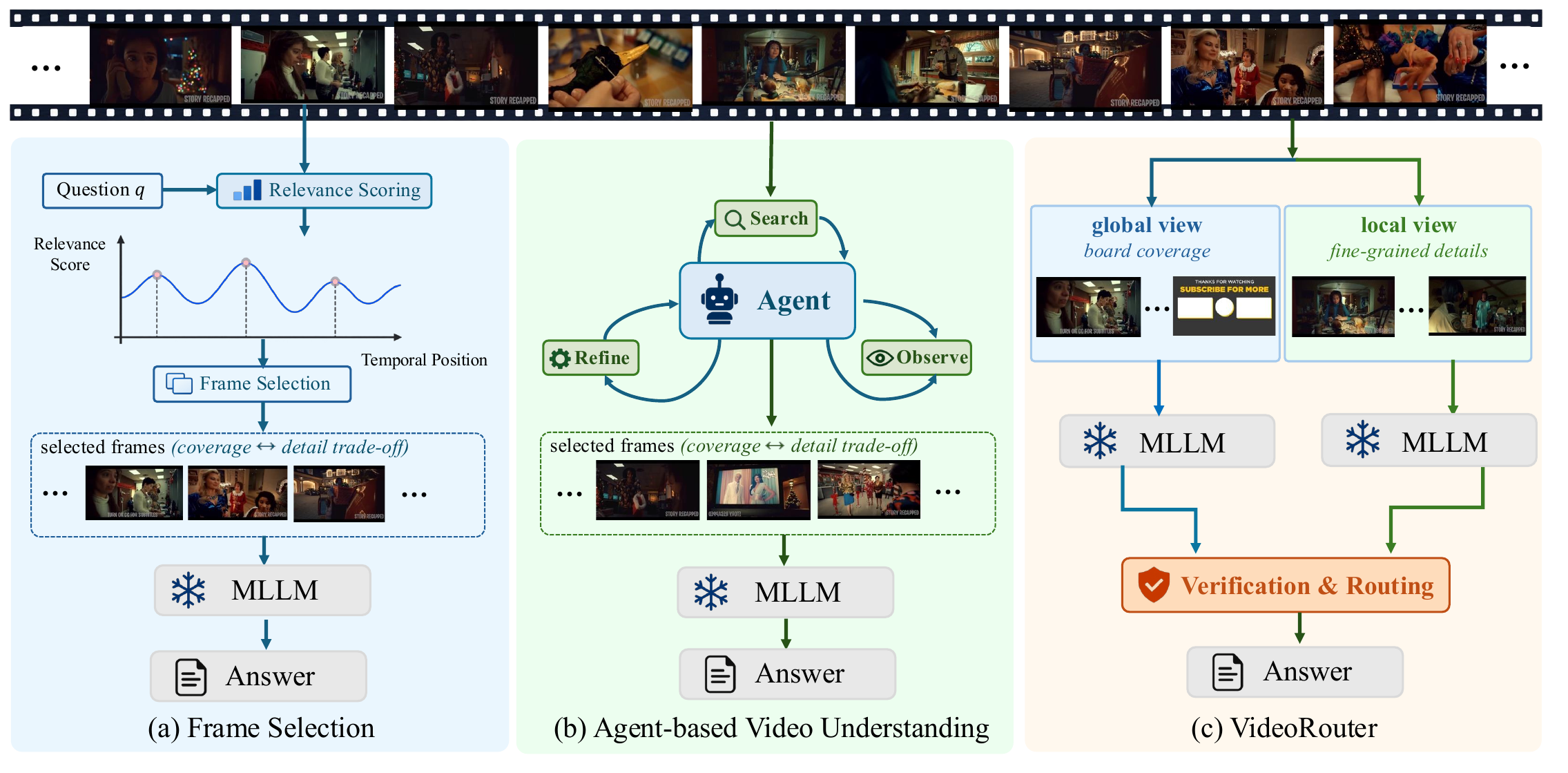}
\caption{Existing frame-selection and agent-based methods ultimately rely on a single frame set, balancing or trading off global coverage and local details. In contrast, VideoRouter maintains separate global and local views and routes between them based on evidence support.}
\label{fig:intro}
\vspace{-1.0em}
\end{figure}


Existing approaches to long-video understanding broadly follow two lines, as illustrated in Figure~\ref{fig:intro}. Frame-selection methods typically estimate frame importance using question-conditioned similarity, and then apply ranking, sampling, or related strategies to balance broad temporal coverage and local relevance in the selected frames~\citep{aks, bolt, qframe, focus}. A second line of work adopts agent-based reasoning, where the model first inspects the video coarsely and then progressively explores question-relevant regions, either retaining broad temporal coverage while adding local evidence or discarding broader information to focus on a local temporal region~\citep{videotree, vca, videorag, videoagent, vgent}. Regardless of how the evidence is selected, both lines ultimately rely on a single frame set to produce the final prediction. However, combining global and local evidence within the same frame set can make decisive evidence less salient, while focusing on a single temporal view can sacrifice either broad coverage or fine-grained details. In this sense, we rethink long-video understanding not as finding one best set of frames, but as coordinating complementary evidence views.

To realize this idea, we need to construct complementary global and local evidence views and determine which view provides stronger support for the final prediction. We propose \textbf{VideoRouter}, a training-free framework that addresses these two needs through two complementary components. First, a shared, question-agnostic temporal hierarchy organizes the video into nested, temporally coherent segments from coarse to fine. From this hierarchy, VideoRouter constructs two complementary views in different ways: a global view selects representative evidence from coarse temporal units to preserve broad coverage, while a local view starts from fine-grained segments and adaptively retains question-relevant regions to recover localized details. Thus, the hierarchy provides a shared search space from which coverage-oriented and detail-oriented evidence can be derived. Second, verification-guided routing evaluates how well each prediction is supported by its corresponding evidence and selects the better-supported prediction. This allows VideoRouter to preserve global and local evidence in separate views and adaptively select the view that provides stronger support for the final prediction, avoiding both evidence dilution from combining them and information loss from relying on a single temporal view only.

Extensive experiments on VideoMME~\citep{videomme} and
LongVideoBench~\citep{longvideobench} demonstrate the effectiveness of our
framework. Verification-guided routing improves over both branches in every
backbone--dataset setting, from 3B up to 27B parameters, and the two views show
opposite dataset-level preferences: VideoMME is better served by broad global
coverage, whereas LongVideoBench relies more on localized evidence. A method
committed to a single fixed view therefore inherits a dataset-dependent bias,
which routing removes. Moreover, on LLaVA-Video-7B, where every baseline gets the same number of MLLM calls as VideoRouter, our method surpasses the strongest frame selection baseline by $2.5$ and $1.3$ points on the two benchmarks respectively.

Our contributions are summarized as follows:
\begin{itemize}
\item We propose VideoRouter, a training-free framework that derives both views from a shared, question-agnostic temporal hierarchy built once per video, and selects between them through evidence-grounded verification.

\item We show that coverage-oriented and relevance-oriented evidence views are strongly complementary for long-video QA, and that the preferred view varies across benchmarks, video durations, and individual questions, so no single fixed view is sufficient.

\item Across six MLLM backbones and two benchmarks, VideoRouter outperforms both individual views in all 12 settings, with gains of up to $+3.89$ on VideoMME and $+4.33$ on LongVideoBench. It also beats the strongest frame selection baseline by $+2.5$ and $+1.3$ points on LLaVA-Video-7B.
\end{itemize}
\section{Related Works}
\subsection{VideoLLM}
Following the success of ChatGPT~\citep{gpt4o}, large language models (LLMs) have opened a new era in artificial intelligence. Beyond advances in natural language processing~\citep{instructgpt, palm}, recent work has extended LLMs to image-language and video-language settings, leading to a series of effective multimodal large language models (MLLMs)~\citep{blip2, mplug, internvl, llavanextvideo, llavaonevision, qwen2, qwen2.5, videollama2, videollava, longva, longvila, timechat, minigpt4, goldfish, sharegpt4video}. Trained on large-scale image-language and video-language data, these models have substantially advanced vision-language understanding. Despite this progress, video MLLMs still face a severe challenge in long-video understanding: their context window remains limited, whereas videos can be far too long to fit within the available token budget. This limitation has motivated a growing body of work on video frame selection and agent-based long-video understanding.

\subsection{Video Frame Selection}
Due to the limited context window of video MLLMs, a line of work has focused on video frame selection~\citep{bolt, aks, qframe, focus}. The goal of these methods is to select a small number of informative frames from a long video, so that the MLLM can answer the question with sufficient visual evidence. AKS~\citep{aks} proposes Adaptive Keyframe Sampling to balance frame--question relevance and temporal coverage over the video. BOLT~\citep{bolt} explores several query-aware frame selection strategies and ultimately adopts inverse transform sampling to select informative frames for long-video understanding. Q-Frame~\citep{qframe} uses the Gumbel-Max trick for efficient frame selection and further adapts the input frame resolution according to similarity scores for better detail understanding.

\subsection{Agent-based Video Understanding}
Recent work has treated MLLMs as agents that iteratively select useful frames and evaluate intermediate predictions, casting long-video understanding as an interactive multi-step reasoning problem~\citep{videoagent, videotree, slowfocus, vca, TTV, longvideor1}. VideoAgent~\citep{videoagent} searches for useful frames in a flat frame pool based on frame positions through multi-round reasoning. VideoTree~\citep{videotree} partitions the video into multiple visually similar clusters and repeatedly queries the MLLM to gather sufficient frames for question answering. VCA~\citep{vca} progressively focuses on selected temporal regions to collect frames and answer the question. Beyond multi-round methods, several one-step approaches have also been proposed for long-video understanding~\citep{videomind, vgent, videomtr, video-r1, vital, videolucy, temporalchainofthought, videorag}. VideoRAG~\citep{videorag} employs the MLLM as a controller to invoke external tools, such as OCR and audio understanding models, for deeper video analysis. VGent~\citep{vgent} first constructs a video graph with semantic relationships for efficient frame retrieval, and then reformulates the question into multiple related sentences to reduce retrieval noise. 

Prior frame-selection and agent-based methods ultimately seek a single question-conditioned visual context, which must trade off broad temporal coverage against fine-grained local relevance. VideoRouter instead constructs complementary global and local views from a shared temporal hierarchy, allowing each to specialize in a different evidence granularity, and uses evidence-grounded verification to route to the better-supported prediction. Our key distinction is therefore adaptive coordination of complementary evidence across temporal scales, rather than single-view selection.

\section{VideoRouter}
\begin{figure}[h]
  \centering
  \includegraphics[width=0.95\linewidth]{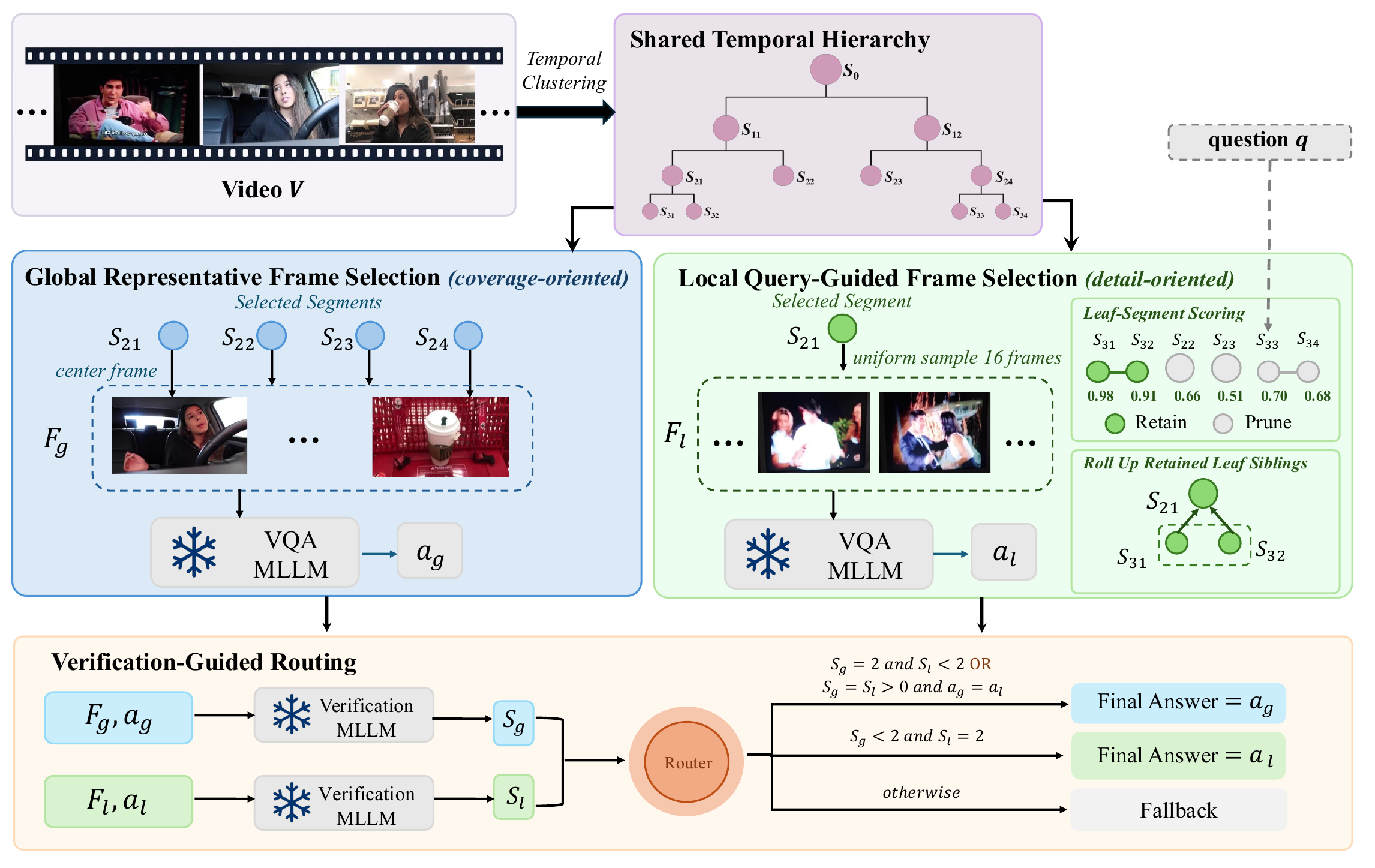}
\caption{\textbf{Overview of the VideoRouter.} The input video is first organized into a temporal hierarchy, where each \(s\) represents a segment containing multiple frames. For clarity, the illustration shows the full video being divided into two coarse segments at the first layer; in practice, however, the first layer typically contains around 32 coarse segments. Based on this hierarchy, we derive a global representative evidence set \(F_g\) and a local query-guided evidence set \(F_l\). The two branches perform VQA independently and are further assessed by verification guided routing.
}
\vspace{-4pt}
\label{fig:framework}
\end{figure}

We consider long-video question answering, where the input consists of a video $V$ and a question $q$, and the goal is to produce an answer $a$ based on the visual evidence in $V$. An overview of VideoRouter is shown in Figure~\ref{fig:framework}.

\subsection{Temporal Hierarchy}
Long-video evidence naturally exists at multiple temporal granularities: some questions require broad storyline coverage, while others depend on fine-grained local details. To support both, we first organize each video into a question-agnostic hierarchical temporal structure that provides coarse-to-fine access to temporally coherent content~\citep{tree}. This hierarchy serves as the common structure for later evidence construction: high-level nodes support global coverage, while low-level nodes preserve detailed visual evidence. In this way, the hierarchy exposes both globally representative frames and fine-grained candidate segments, enabling the later branches to access evidence at the temporal scale required by the question.

Given a video \(V=\{v_1,\ldots,v_t\}\), we first extract a visual feature $f_i$ for each sampled frame using a visual encoder $\phi(\cdot)$:
\[
f_i = \phi(v_i) \in \mathbb{R}^d, \quad i=1,\ldots,t.
\]

We then build the temporal hierarchy using contiguous temporal clustering~(Pseudo Code is in Appendix~\ref{sec:pseudo_code}). At the first level, we partition the video into \(M\) coarse temporal segments. This coarse partition is designed to preserve broad temporal coverage, and in practice often roughly aligns with major scene or story transitions. Starting from these coarse segments, we recursively split each non-leaf segment into temporally contiguous child segments, producing a coarse-to-fine hierarchy. The recursion stops when the segment length falls below a predefined threshold \(L_{\text{leaf}}\)=16, so that deeper nodes correspond to increasingly localized temporal details~(Details are in Appendix~\ref{sec:clustering}). 
The recursive construction induces a nested coarse-to-fine temporal partition. 
For each non-leaf segment \(s\), its child segments 
\(\mathcal{C}(s)=\{c_1,\ldots,c_K\}\) are contiguous and non-overlapping, and jointly cover the parent segment:
\[
s=\bigcup_{c\in\mathcal{C}(s)} c,
\qquad
c_i\cap c_j=\varnothing~(i\neq j),
\qquad
\sum_{c\in\mathcal{C}(s)} |c| = |s|.
\]
Since each child satisfies the minimum-length constraint \(L_{\min}\), its temporal span is bounded by
\[
L_{\min} \le |c| \le |s|-(K-1)L_{\min}.
\]
Consequently, temporal support progressively decreases along each root-to-leaf path, providing increasingly fine-grained access to the video while preserving the temporal coverage of its ancestors.




For each \(s\), we additionally define a representative center frame, which is later used by the representative branch. Specifically, we compute the mean feature of frames in segment and select the frame with the highest cosine similarity to it:
\[
\bar{f}_s=\frac{1}{|s|}\sum_{i\in s} f_i, \qquad
c(s)=\arg\max_{i \in s}\cos(f_i,\bar{f}_s).
\]
This center frame provides a compact visual anchor for the segments.

\subsection{Global Representative Reasoning}
We construct a global evidence view that preserves broad temporal coverage while representing each temporal unit compactly. Instead of uniformly sampling frames from the entire video, we select representative frames from a fixed level of the temporal hierarchy.

Specifically, let the first coarse partition be level \(1\). Given a representation depth \(d_{\text{rep}}\), we traverse each branch of the hierarchy and stop when reaching level \(d_{\text{rep}}\), or at the deepest available node if the branch terminates earlier. 
As illustrated in Figure~\ref{fig:framework}, a smaller \(d_{\text{rep}}\) yields coarser temporal units, while a larger \(d_{\text{rep}}\) provides finer temporal coverage. We take the center frame \(c(s)\) of each selected segment and retain these frames in temporal order to form the global evidence \(F_g\).
Because the selected segments jointly span the video, \(F_g\) provides a compact, coverage-oriented representation of its overall temporal structure. The global branch then produces an answer using the selected visual evidence and the question:
\[
a_g = \mathrm{MLLM}(F_g, q).
\]



\subsection{Local Query-Guided Reasoning}
To recover fine-grained evidence that may be missed by the global branch, we construct a local evidence view by pruning the temporal hierarchy toward question-relevant regions. Specifically, we start from the deepest available segments along each root-to-leaf path, which provide the finest-grained temporal candidates in the hierarchy. Our goal is to retain only those temporally coherent segments that contain strong evidence for the current question, while discarding weak or irrelevant branches, as illustrated in Figure~\ref{fig:framework}.

Given the question feature \(f_q\), we assign each segment \(s\) a relevance score equal to the highest frame-level relevance within that segment:
\(
r(s)=\max_{i\in s}\cos(f_i,f_q).
\)
This max-based scoring is suitable for long-video QA, where decisive evidence may be temporally sparse and appear in only a small portion of a segment. We then define a pruning threshold to decide which segments should be preserved. Specifically, let
$
r_{\max}=\max_s r(s)
$
be the highest relevance score among all candidate segments. We set
$
\tau=\alpha r_{\max},
$
where \(\alpha\in(0,1]\) is the prune ratio. Intuitively, this keeps only segments whose relevance is sufficiently close to the strongest candidate. Segments with \(r(s)<\tau\) are removed, while segments with \(r(s)\ge\tau\) are preserved as local evidence candidates. This relative threshold makes the pruning adaptive to each question, rather than relying on a fixed absolute similarity value. After pruning, we further reduce fragmentation through a roll-up operation. If all original children(before pruning) of a parent segment are retained, we replace them with the parent segment, whose retained score is defined as
$
r'(p)=\max_{c\in\operatorname{Child}(p)} r'(c).
$
This preserves the strongest relevance signal while recovering a more coherent temporal unit.

Finally, from each retained segment, we uniformly sample 16 frames to form a local evidence set \(F_l\). The local prediction is then obtained as
\[
a_l=\mathrm{MLLM}(F_l, q).
\]


\subsection{Verification-Guided Routing}
The global and local branches provide complementary evidence views, but their predictions may not be equally reliable for a given question. To resolve this uncertainty, we explicitly assess whether each branch prediction is supported by its corresponding visual evidence.

Specifically, for each branch \(b\in\{g,l\}\), we perform a separate verification pass using an MLLM conditioned on the selected visual evidence \(F_b\), the question \(q\), and the branch prediction \(a_b\):
\[
s_b=\mathrm{Verify}(F_b,q,a_b), \qquad s_b\in\{0,1,2\}.
\]
The verifier evaluates how strongly the selected visual evidence supports each branch prediction using three discrete levels:

\begin{itemize}
    \item \textit{2 (Strong): Clear and sufficient visual evidence.}
    \item \textit{1 (Partial): Relevant but incomplete or ambiguous evidence.}
    \item \textit{0 (None): No relevant supporting visual evidence.}
\end{itemize}
The detailed verification prompt is provided in Appendix~\ref{sec:prompts}.

We then route between the two branch predictions according to their verification scores. When exactly one branch receives strong support, i.e., \((s_g,s_l)\in\{(2,1),(2,0)\}\) or \((1,2),(0,2)\}\), we directly accept the prediction from the branch with score \(2\). When both branches receive the same non-zero score and produce the same answer, we also accept their shared prediction. This covers both \((2,2)\), where the two views independently provide strong support for the same answer, and \((1,1)\), where neither view is individually conclusive but their predictions are consistent. Formally, the routing decision is
\[
a_r =
\begin{cases}
a_g, & s_g=2,\; s_l<2,\\
a_l, & s_l=2,\; s_g<2,\\
a_g, & s_g=s_l>0 \text{ and } a_g=a_l,\\
\mathrm{Unresolved}, & \text{otherwise}.
\end{cases}
\]

For these unresolved cases, we combine the complementary evidence from the two branches and perform an additional reasoning pass. Let
\(
F_u = F_g \cup F_l
\)
denote the union of the global and local evidence. The final prediction is then obtained as
\[
a_u=\mathrm{MLLM}(F_u,q).
\]
This fallback allows the model to jointly reconsider broad temporal context and fine-grained local evidence when verification cannot confidently resolve the two branches.

\section{Experiments}
\subsection{Experiments Setups}
\paragraph{Implementation Details.}
For offline temporal hierarchy construction, each video is sampled at 2.0 FPS, with frame features extracted using the CLIP vision encoder and question features obtained from the CLIP text encoder~\citep{clip} for query-guided relevance scoring. In the online stage, the global representative branch uses \(d_{\mathrm{rep}}=2\), resulting in 32--64 representative center frames per video, while the local query-guided branch uses \(\alpha=0.95\) as the pruning threshold. The visual input is capped at 128 frames for Qwen-based models and 64 frames for LLaVA-Video, with uniform subsampling applied whenever the selected frames exceed these limits. Unless otherwise specified, the QA and verification modules share the same MLLM backbone. For controlled evaluation, experiments focus on multiple-choice QA using VideoMME~\citep{videomme} and LongVideoBench~\citep{longvideobench}. All experiments are conducted on NVIDIA A100 80GB GPUs.

\subsection{Main Results}
\begin{table}[h]
\definecolor{lightblue}{RGB}{235,242,250}
\definecolor{lightgrey}{RGB}{235,235,235}
\centering
\small
\setlength{\tabcolsep}{6.0pt}
\renewcommand{\arraystretch}{1.12}
\caption{\textbf{Comparison of global, local, and verification-guided routing strategies on VideoMME and LongVideoBench.} Oracle denotes the upper-bound accuracy where a sample is considered correct if either the global or local branch predicts the correct answer. Red numbers denote gains over the stronger single view (Local or Global).}
\begin{tabular}{l|l|ccc|c}
\toprule
\multirow{2}{*}{\textbf{Dataset}} &
\multirow{2}{*}{\textbf{Model}} &
\multicolumn{4}{c}{\textbf{Method}} \\
\cmidrule(lr){3-6}
& & \textbf{Local} & \textbf{Global} & \textbf{Router} & \textbf{Oracle} \\
\midrule

\multirow{6}{*}{\textbf{VideoMME}}
& Qwen2.5-VL-3B~\citep{qwen2.5}
& 56.37 & 57.85
& \cellcolor{lightblue}\(60.00_{\scriptscriptstyle\red{+2.15}}\)
& 67.67 \\

& Qwen3-VL-4B~\citep{qwen3}
& 58.37 & 63.26
& \cellcolor{lightblue}\(65.85_{\scriptscriptstyle\red{+2.59}}\) & 71.96 \\ 

& Qwen3.5-VL-4B~\citep{qwen35}
& 60.65 & 64.60
& \cellcolor{lightblue}\(67.53_{\scriptscriptstyle\red{+2.93}}\)
& 73.25 \\

& Qwen2-VL-7B~\citep{qwen2}
& 58.30 & 58.41
& \cellcolor{lightblue}\(61.26_{\scriptscriptstyle\red{+2.85}}\)
& 68.41 \\

& Qwen2.5-VL-7B~\citep{qwen2.5}
& 60.11 & 61.63
& \cellcolor{lightblue}\(65.52_{\scriptscriptstyle\red{+3.89}}\)
& 71.63 \\

& LLaVA-Video-7B~\citep{llavanextvideo}
& 59.33 & 64.11
& \cellcolor{lightblue}\(64.37_{\scriptscriptstyle\red{+0.26}}\)
& 71.93 \\

\midrule

\multirow{6}{*}{\textbf{LVB}}
& Qwen2.5-VL-3B~\citep{qwen2.5}
& 55.07 & 53.24
& \cellcolor{lightblue}\(58.40_{\scriptscriptstyle\red{+3.33}}\)
& 64.59 \\

& Qwen3-VL-4B~\citep{qwen3}
& 61.06 & 58.82
& \cellcolor{lightblue}\(65.39_{\scriptscriptstyle\red{+4.33}}\)
& 72.80 \\

& Qwen3.5-VL-4B~\citep{qwen35}
& 58.90 & 56.29
& \cellcolor{lightblue}\(63.14_{\scriptscriptstyle\red{+4.24}}\)
& 69.05 \\

& Qwen2-VL-7B~\citep{qwen2}
& 56.16 & 52.25
& \cellcolor{lightblue}\(57.65_{\scriptscriptstyle\red{+1.49}}\)
& 65.47 \\

& Qwen2.5-VL-7B~\citep{qwen2.5}
& 60.32 & 57.32
& \cellcolor{lightblue}\(64.23_{\scriptscriptstyle\red{+3.91}}\)
& 69.88 \\

& LLaVA-Video-7B~\citep{llavanextvideo}
& 60.48 & 59.32
& \cellcolor{lightblue}\(63.56_{\scriptscriptstyle\red{+3.08}}\)
& 71.05 \\

\bottomrule
\end{tabular}
\label{tab:main_result}
\end{table}

\paragraph{Global, Local, and Routed Reasoning.}
Table~\ref{tab:main_result} reports the local branch, the global branch, the
verification-guided router, and the oracle across six backbones on VideoMME and
LongVideoBench. Two trends emerge: 

\textit{The two views exhibit opposite, dataset-level preferences.}
On VideoMME, the global branch outperforms the local branch for all six
backbones, by margins ranging from $+0.11$ (Qwen2-VL-7B) to $+4.89$
(Qwen3-VL-4B), indicating that this benchmark is largely resolved by broad
temporal coverage and high-level storyline context. On LVB the ordering reverses
just as consistently: the local branch wins on all six backbones, by $+1.16$ to
$+3.91$ points. This flip is systematic rather than backbone-specific, and it
shows that no single evidence granularity is universally preferable. A method
committed to one fixed view therefore inherits a dataset-dependent bias, whereas
the appropriate granularity has to be decided per benchmark, and in fact per
question.

\textit{Verification-guided routing turns this complementarity into accuracy.}
Whichever branch is stronger in isolation, the oracle exceeds the better of the two by $7.82$--$11.74$ points, so the two views are genuinely complementary rather than competing approximations of the same evidence. The router improves over both branches in all 12 settings: measured against the stronger branch, it gains $+0.26$ to $+3.89$ points on VideoMME and $+1.49$ to $+4.33$ points on LVB. The gains are larger on LVB, where fine-grained evidence is decisive and the two views disagree on more samples.Notably, this advantage is independent of model capacity: it holds for every backbone we test, from 3B up to 27B (Appendix~\ref{sec:additional_backbones}), and does not
shrink as the backbone grows.


\begin{table}[h]
\centering
\small
\definecolor{lightblue}{RGB}{235,242,250}
\begin{minipage}[t]{0.50\linewidth}
\centering
\setlength{\tabcolsep}{2pt}
\renewcommand{\arraystretch}{1.12}
\caption{Comparison with state-of-the-art video agents and representative proprietary MLLMs. VideoRouter-L based on LLaVA-Video-7B, while VideoRouter-Q based on Qwen2.5-7B}
\begin{tabular}{lccc}
\toprule
\textbf{Model} & \textbf{Size} & \textbf{VideoMME} & \textbf{LVB} \\
\midrule
\multicolumn{4}{l}{\textbf{\textit{Proprietary MLLMs}}} \\
GPT-4o~\citep{gpt4o}        & -  & 72.1 & 66.7 \\
Gemini 1.5 Pro~\citep{gemini} & -  & 75.7 & 62.9 \\
\midrule
\multicolumn{4}{l}{\textbf{\textit{Video Agents}}} \\
VideoTree~\citep{videotree}            & 7B  & 56.1    & 52.3 \\
VideoRAG~\citep{videorag}       & 7B & 62.1 & 58.7 \\
VideoMind~\citep{videomind}      & 7B & 58.2 & 56.3 \\
\midrule
\rowcolor{lightblue}
VideoRouter-L         & 7B & 64.4 & 63.6 \\
\rowcolor{lightblue}
VideoRouter-Q         & 7B & 65.5 & 64.2 \\
\bottomrule
\end{tabular}
\label{tab:soa_compare}
\end{minipage}
\hfill
\begin{minipage}[t]{0.45\linewidth}
\centering
\setlength{\tabcolsep}{1pt}
\renewcommand{\arraystretch}{1.12}
\caption{Comparison with state-of-the-art adaptive frame selection methods.
Every baseline uses 5 runs with majority voting, matching VideoRouter's maximum number of MLLM calls in a single run.}
\begin{tabular}{lc}
\toprule
\textbf{Model}  & \textbf{LVB} \\
\midrule
LLaVA-Video-7B~\citep{llavanextvideo}  & 56.7 \\
~~\textit{w/} AKS~\citep{aks}      & 62.3 \\
~~\textit{w/} Q-Frame~\citep{qframe}  & 60.1 \\
\rowcolor{lightblue}
~~\textit{w/} VideoRouter     & 63.6 \\
\bottomrule
\bottomrule
\textbf{Model}  & \textbf{VideoMME} \\
\midrule
LLaVA-Video-7B~\citep{llavanextvideo}  &  60.4\\
~~\textit{w/} AKS~\citep{aks}      &  61.9\\
\rowcolor{lightblue}
~~\textit{w/} VideoRouter     &  64.4\\
\bottomrule
\end{tabular}
\label{tab:lvb_llava_compare}
\end{minipage}
\end{table}

\paragraph{Comparison with Existing Video Agents.}
Table~\ref{tab:soa_compare} compares our method with proprietary MLLMs and prior video agents on VideoMME and LVB. Among open 7B-scale methods, both of our variants achieve the strongest results on the two benchmarks. In particular, VideoRouter-Q reaches 65.5 on VideoMME and 64.2 on LVB, outperforming VideoRAG by +3.4 and +5.5 points, respectively. The gains over VideoMind are even larger, reaching +7.3 on VideoMME and +7.9 on LVB. These results show that our global--local evidence construction with verification-guided routing is substantially more effective than prior video agents.

\paragraph{Comparison with Frame Selection Methods.}
Table~\ref{tab:lvb_llava_compare} compares VideoRouter with adaptive frame
selection methods on  LLaVA-Video-7B backbone, with all methods seeing
at most 64 frames per VQA call. A single VideoRouter run issues at most 5 MLLM
calls, of which 3 are VQA and 2 are verification. For a fair comparison, we grant each baseline the same
call budget: 5 independent full 64-frame VQA runs at
temperature $0.7$, aggregated by majority voting. VideoRouter is nonetheless the
strongest method on both benchmarks, reaching $64.4$ on VideoMME and $63.6$ on
LVB. It improves over the LLaVA-Video-7B baseline by $+4.0$ and $+6.9$ points,
and over the best frame selection method, AKS, by $+2.5$ and $+1.3$ points. AKS
and Q-Frame already improve substantially over uniform sampling, confirming that
frame selection matters. These results indicate that the improvement does not come merely from selecting better individual frames, but from organizing them into complementary global and local evidence views and resolving them through routing.

\subsection{Ablation Studies}
\begin{figure}[h]
  \centering
  \includegraphics[width=0.85\linewidth]{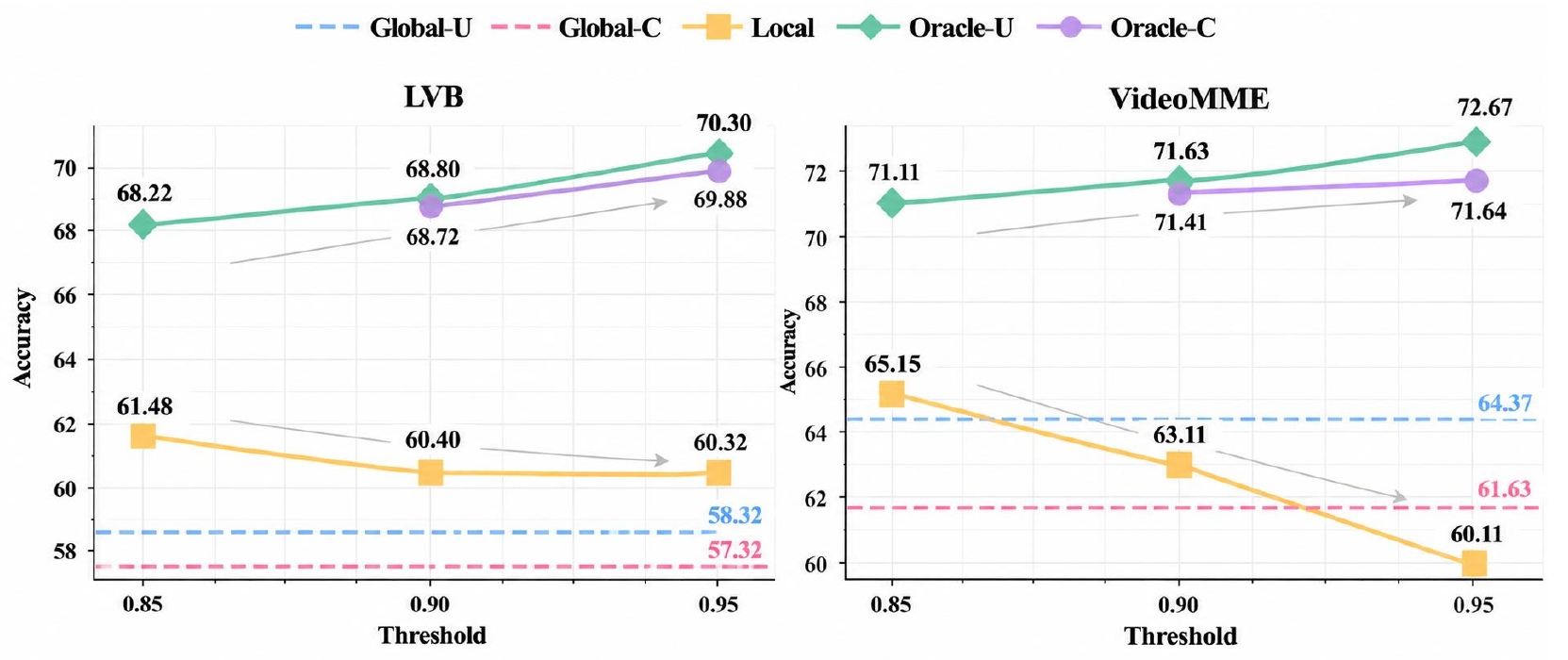}
\caption{\textbf{Effect of local pruning on global--local complementarity.} For the global branch, Global-U uniformly samples 4 frames from each segment at \(d_{\mathrm{rep}}=2\), while Global-C uses the center frame of each segment at the same level.
Increasing the pruning threshold \(\alpha\) weakens the standalone local branch but consistently improves the oracle accuracy when paired with the fixed global branch, indicating stronger complementarity.}
\vspace{-8pt}
\label{fig:thre_oracle}
\end{figure}

\paragraph{Complementarity from the Temporal Hierarchy.}
Figure~\ref{fig:thre_oracle} studies whether the temporal hierarchy can produce complementary global and local evidence views. The global branch is kept fixed, while increasing \(\alpha\) makes the local branch increasingly selective by retaining only segments with stronger question relevance. Interestingly, this makes the local branch weaker in isolation but more complementary to the global view. On LVB, the local accuracy decreases from \(61.48\) at \(\alpha=0.85\) to \(60.32\) at \(\alpha=0.95\), whereas Oracle-U increases from \(68.22\) to \(70.30\). Accordingly, the oracle gain over the stronger individual branch grows from \(6.74\) to \(9.98\) points. A similar pattern appears on VideoMME: local accuracy drops from \(65.15\) to \(60.11\), while Oracle-U rises from \(71.11\) to \(72.67\), increasing the oracle gain over the stronger single branch from \(5.96\) to \(8.30\) points. Oracle-C follows the same increasing trend, showing that the observation is consistent across different global representations. These results show that the temporal hierarchy provides a shared structure for constructing complementary global and local evidence views at different temporal granularities. As local pruning becomes stricter, the local view becomes more specialized and less redundant with the global view, leading to a larger oracle gap despite lower standalone local accuracy.

\begin{figure}[h]
  \centering
  \includegraphics[width=0.85\linewidth]{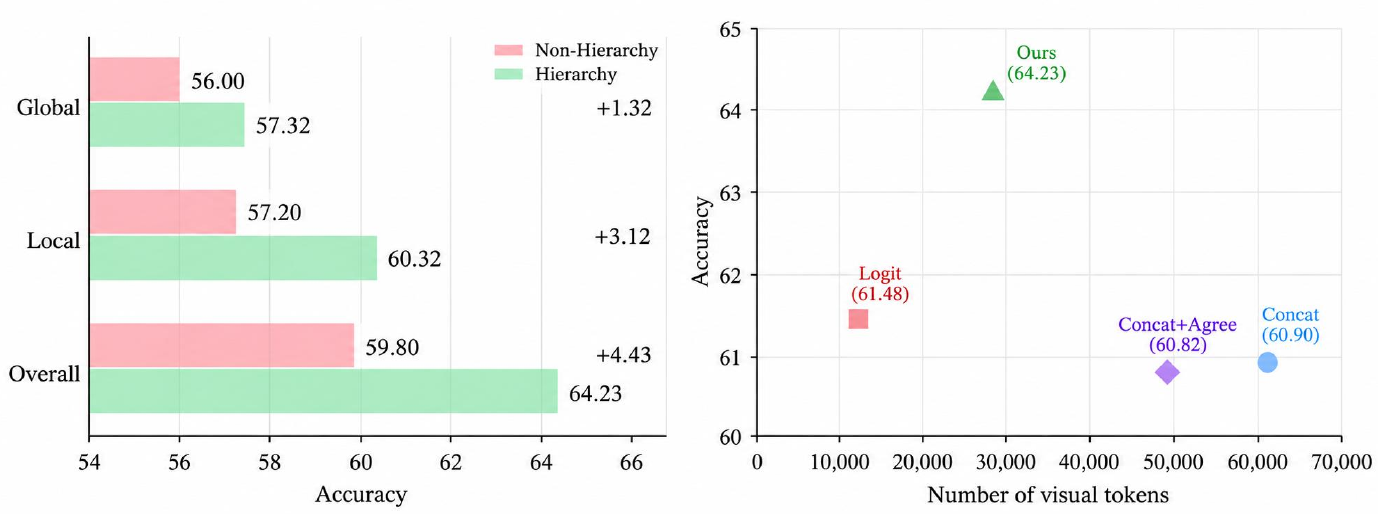}
\caption{Comparison on LongVideoBench with Qwen2.5-VL-7B.
Left: ablation comparing No-Hierarchy Router and VideoRouter under the same verification-guided routing pipeline. Right: lightweight ensemble methods using the same global and local evidence views.}
\vspace{-8pt}
\label{fig:abla}
\end{figure}

\paragraph{No-Hierarchy Routing.}
We added a controlled ablation that removes the temporal hierarchy in Figure~\ref{fig:abla}. The global branch uniformly samples 64 frames because the corresponding hierarchy level contains approximately 64 segment representatives; this matches the global evidence scale while isolating the effect of hierarchical organization. The local branch directly retrieves frames from a flat pool using $s_i \geq 0.95s_{\max}$. Both branches then use the same verification and fallback pipeline as VideoRouter. This experiment directly covers uniform sampling, flat CLIP retrieval, and routing without hierarchical construction. Under the same routing pipeline, VideoRouter improves over the no-hierarchy router by $4.43$ points, demonstrating the importance of hierarchy-based complementary evidence construction. 

\paragraph{Comparison with Lightweight Ensemble Methods.}
We further compare VideoRouter with lightweight alternatives built on the same global and local evidence views. Specifically, we consider:

(1) \emph{Next-token confidence}, which selects the branch whose predicted answer has the higher next-token probability, following prior work that uses next-token probabilities as a proxy for model confidence~\citep{con1, con2};

(2) \emph{Concat}, which directly reasons over the union of the global and local evidence. To roughly match the number of MLLM calls used by VideoRouter, we perform five inference runs with temperature \(0.7\) and aggregate the predictions by majority voting;

(3) \emph{Agree+Concat}, which directly accepts the prediction when the global and local branches agree and otherwise falls back to reasoning over their concatenated evidence. Since the two branch predictions already require two MLLM calls, we perform three additional inference runs with temperature \(0.7\) for the fallback cases and use majority voting, yielding a comparable maximum number of MLLM calls as VideoRouter.

As shown in Figure~\ref{fig:abla}, next-token confidence reaches \(61.48\), remaining \(2.75\) points below VideoRouter. Concat and Agreement+Concat achieve \(60.90\) and \(60.82\), respectively, while consuming substantially more visual tokens than VideoRouter, which reaches \(64.23\). These results show that the improvement does not simply come from processing more visual tokens. Instead, explicitly verifying whether each prediction is supported by its corresponding visual evidence enables more effective use of the available visual evidence than output confidence, evidence concatenation, or branch agreement alone.
\section{Conclusion}
In this paper, we argue that long-video understanding should not be viewed as selecting a single best subset of frames, but as coordinating complementary evidence across different temporal scales. Based on this perspective, we propose VideoRouter, a training-free framework that organizes videos into a question-agnostic temporal hierarchy. This hierarchy supports a global reasoning view for capturing broad storyline context and event progression, as well as a local reasoning view for recovering fine-grained, question-relevant details. VideoRouter further introduces verification-guided routing to evaluate how well each prediction is supported by its visual evidence and to adaptively select or combine the two views. Extensive experiments on VideoMME and LongVideoBench show consistent improvements over single-branch reasoning, frame-selection methods, and prior video agents. These results demonstrate that effective long-video understanding depends not only on finding informative frames, but also on organizing, verifying, and coordinating complementary evidence. We hope this perspective can inspire more reliable and flexible long-video reasoning methods for MLLMs.

\subsection*{AI use statement}
(This section is \textbf{required} and does not count toward the page limit.)

In this work, generative AI tools were used only to assist with writing, such as correcting grammar and improving wording and readability. They were not used for any substantive research tasks, including ideation, method or experiment design, implementation, data processing, result analysis, or drawing conclusions, all of which were carried out by the authors. All AI-assisted text was carefully reviewed and revised by the authors, who take full responsibility for the content of this paper.

\bibliography{iclr2027_conference}

@String(CVPR= {IEEE Conf. Comput. Vis. Pattern Recog.})

@String(ICCV= {Int. Conf. Comput. Vis.})

@String(ECCV= {Eur. Conf. Comput. Vis.})

@String(NIPS= {Adv. Neural Inform. Process. Syst.})

@String(EMNLP  = {EMNLP})

@String(CVPR  = {CVPR})

@String(ICCV  = {ICCV})

@String(ECCV  = {ECCV})

@String(NIPS  = {NeurIPS})

@String(ICML = {ICML})

@inproceedings{tree,
  title={Category-specific video summarization},
  author={Potapov, Danila and Douze, Matthijs and Harchaoui, Zaid and Schmid, Cordelia},
  booktitle=ECCV,
  year={2014},
}

@inproceedings{longvideor1,
    author    = {Qiu, Jihao and Xie, Lingxi and Huo, Xinyue and Tian, Qi and Ye, Qixiang},
    title     = {LongVideo-R1: Smart Navigation for Low-cost Long Video Understanding},
    booktitle = CVPR,
    year      = {2026},
}

@inproceedings{TTV,
    author    = {Wang, Zheng and Chen, Haoran and Qin, Haoxuan and Wei, Zhipeng and Qian, Tianwen and Bai, Cong},
    title     = {Think, Then Verify: A Hypothesis-Verification Multi-Agent Framework for Long Video Understanding},
    booktitle = CVPR,
    year      = {2026},
}

@inproceedings{bolt,
    title     = {BOLT: Boost Large Vision-Language Model Without Training for Long-form Video Understanding},
    author    = {Liu, Shuming and Zhao, Chen and Xu, Tianqi and Ghanem, Bernard},
    booktitle = CVPR,
    year      = {2025},
}

@article{qwen3,
  title={Qwen3-vl technical report},
  author={Bai, Shuai and Cai, Yuxuan and Chen, Ruizhe and Chen, Keqin and Chen, Xionghui and Cheng, Zesen and Deng, Lianghao and Ding, Wei and Gao, Chang and Ge, Chunjiang and others},
  journal={arXiv:2511.21631},
  year={2025}
}

@misc{qwen35,
    title = {Qwen3.5: Towards Native Multimodal Agents},
    url = {https://qwen.ai/blog?id=qwen3.5},
    author = {Qwen Team},
    year = {2026}
}

@inproceedings{con2,
  title={Dyfo: A training-free dynamic focus visual search for enhancing lmms in fine-grained visual understanding},
  author={Li, Geng and Xu, Jinglin and Zhao, Yunzhen and Peng, Yuxin},
  booktitle=CVPR,
  year={2025}
}

@inproceedings{con1,
  title={Commonsense video question answering through video-grounded entailment tree reasoning},
  author={Liu, Huabin and Ilievski, Filip and Snoek, Cees GM},
  booktitle=CVPR, 
  year={2025}
}

@article{gemini,
  title={Gemini 1.5: Unlocking multimodal understanding across millions of tokens of context},
   author= {Gemini Team Google},
  journal={arXiv:2403.05530},
  year={2024}
}

@article{temporalchainofthought,
  title={Temporal Chain of Thought: Long-Video Understanding by Thinking in Frames},
  author={Arnab, Anurag and Iscen, Ahmet and Caron, Mathilde and Fathi, Alireza and Schmid, Cordelia},
  journal={arXiv:2507.02001},
  year={2025}
}

@article{videolucy,
  title={VideoLucy: Deep Memory Backtracking for Long Video Understanding},
  author={Zuo, Jialong and Deng, Yongtai and Kong, Lingdong and Yang, Jingkang and Jin, Rui and Zhang, Yiwei and Sang, Nong and Pan, Liang and Liu, Ziwei and Gao, Changxin},
  journal={arXiv:2510.12422},
  year={2025}
}

@article{vital,
  title={Thinking with videos: Multimodal tool-augmented reinforcement learning for long video reasoning},
  author={Zhang, Haoji and Gu, Xin and Li, Jiawen and Ma, Chixiang and Bai, Sule and Zhang, Chubin and Zhang, Bowen and Zhou, Zhichao and He, Dongliang and Tang, Yansong},
  journal={arXiv:2508.04416},
  year={2025}
}

@article{videomtr,
  title={Video-mtr: Reinforced multi-turn reasoning for long video understanding},
  author={Xie, Yuan and Chen, Tianshui and Ge, Zheng and Ni, Lionel},
  journal={arXiv:2508.20478},
  year={2025}
}

@article{videorag,
  title={Video-rag: Visually-aligned retrieval-augmented long video comprehension},
  author={Luo, Yongdong and Zheng, Xiawu and Li, Guilin and Yin, Shukang and Lin, Haojia and Fu, Chaoyou and Huang, Jinfa and Ji, Jiayi and Chao, Fei and Luo, Jiebo and Rongrong Ji},
  journal={arXiv:2411.13093},
  year={2024}
}

@inproceedings{slowfocus,
  title={Slowfocus: Enhancing fine-grained temporal understanding in video llm},
  author={Nie, Ming and Ding, Dan and Wang, Chunwei and Guo, Yuanfan and Han, Jianhua and Xu, Hang and Zhang, Li},
  booktitle = NIPS,
  year={2024}
}

@article{gpt4o,
  title={Gpt-4o system card},
  author={Hurst, Aaron and Lerer, Adam and Goucher, Adam P and Perelman, Adam and Ramesh, Aditya and Clark, Aidan and Ostrow, AJ and Welihinda, Akila and Hayes, Alan and Radford, Alec and others},
  journal={arXiv:2410.21276},
  year={2024}
}

@inproceedings{vca,
  title={Vca: Video curious agent for long video understanding},
  author={Yang, Zeyuan and Chen, Delin and Yu, Xueyang and Shen, Maohao and Gan, Chuang},
  booktitle=ICCV,
  year={2025}
}

@article{llavaonevision,
  	title={LLaVA-OneVision: Easy Visual Task Transfer},
  	author={Li, Bo and Zhang, Yuanhan and Guo, Dong and Zhang, Renrui and Li, Feng and Zhang, Hao and Zhang, Kaichen and Li, Yanwei and Liu, Ziwei and Li, Chunyuan},
  	journal={arXiv:2408.03326},
  	year={2024}
}

@article{longva,
  title={Long Context Transfer from Language to Vision},
  author={Zhang, Peiyuan and Zhang, Kaichen and Li, Bo and Zeng, Guangtao and Yang, Jingkang and Zhang, Yuanhan and Wang, Ziyue and Tan, Haoran and Li, Chunyuan and Liu, Ziwei},
  journal={arXiv:2406.16852},
  year={2024},
}

@inproceedings{vgent,
    title={Vgent: Graph-based Retrieval-Reasoning-Augmented Generation For Long Video Understanding},
    author={Shen, Xiaoqian and Zhang, Wenxuan and Chen, Jun and Elhoseiny, Mohamed},
    booktitle = NIPS,
    year={2025}
  }

@inproceedings{videomme,
  title={Video-MME: The First-Ever Comprehensive Evaluation Benchmark of Multi-modal LLMs in Video Analysis},
  author={Fu, Chaoyou and Dai, Yuhan and Luo, Yongdong and Li, Lei and Ren, Shuhuai and Zhang, Renrui and Wang, Zihan and Zhou, Chenyu and Shen, Yunhang and Zhang, Mengdan and others},
  booktitle=CVPR,
  year={2025}
}

@inproceedings{longvideobench,
        title={LongVideoBench: A Benchmark for Long-context Interleaved Video-Language Understanding}, 
        author={Wu, Haoning and Li, Dongxu and Chen, Bei and Li, Junnan},
        booktitle=NIPS,
        year={2024},
  }

@inproceedings{videollava,
  title={Video-llava: Learning united visual representation by alignment before projection},
  author={Lin, Bin and Ye, Yang and Zhu, Bin and Cui, Jiaxi and Ning, Munan and Jin, Peng and Yuan, Li},
  booktitle=EMNLP,
  year={2024}
}

@inproceedings{sharegpt4video,
  title={Sharegpt4video: Improving video understanding and generation with better captions},
  author={Chen, Lin and Wei, Xilin and Li, Jinsong and Dong, Xiaoyi and Zhang, Pan and Zang, Yuhang and Chen, Zehui and Duan, Haodong and Bin, Lin and Tang, Zhenyu and Yuan, Li and Yu, Qiao and Lin, Dahua and Zhao, Feng and Wang, Jiaqi},
  booktitle=NIPS,
  year={2024}
}

@article{llavanextvideo,
    title={Video Instruction Tuning With Synthetic Data}, 
    author={Yuanhan Zhang and Jinming Wu and Wei Li and Bo Li and Zejun Ma and Ziwei Liu and Chunyuan Li},
    journal={arXiv:2410.02713},
    year={2024}
}

@article{videollama2,
  title={VideoLLaMA 2: Advancing Spatial-Temporal Modeling and Audio Understanding in Video-LLMs},
  author={Cheng, Zesen and Leng, Sicong and Zhang, Hang and Xin, Yifei and Li, Xin and Chen, Guanzheng and Zhu, Yongxin and Zhang, Wenqi and Luo, Ziyang and Zhao, Deli and Bing, Lidong},
  journal={arXiv:2406.07476},
  year={2024},
}

@article{qwen2.5,
  title={Qwen2. 5-vl technical report},
  author={Bai, Shuai and Chen, Keqin and Liu, Xuejing and Wang, Jialin and Ge, Wenbin and Song, Sibo and Dang, Kai and Wang, Peng and Wang, Shijie and Tang, Jun and Zhong, Humen and Zhu, Yuanzhi and Yang, Mingkun and Li, Zhaohai and Wan, Jianqiang and Wang, Pengfei and Ding, Wei and Fu, Zheren and Xu, Yiheng and Ye, Jiabo and Zhang, Xi and Xie, Tianbao and Cheng, Zesen and Zhang, Hang and Yang, Zhibo and Xu, Haiyang and Lin, Junyang},
  journal={arXiv:2502.13923},
  year={2025}
}

@inproceedings{videotree,
  title={Videotree: Adaptive tree-based video representation for llm reasoning on long videos},
  author={Wang, Ziyang and Yu, Shoubin and Stengel-Eskin, Elias and Yoon, Jaehong and Cheng, Feng and Bertasius, Gedas and Bansal, Mohit},
  booktitle=CVPR,
  year={2025}
}

@article{videomind,
  title={VideoMind: A Chain-of-LoRA Agent for Long Video Reasoning},
  author={Liu, Ye and Lin, Kevin Qinghong and Chen, Chang Wen and Shou, Mike Zheng},
  journal={arXiv:2503.13444},
  year={2025}
}

@inproceedings{video-r1,
  title={Video-R1: Reinforcing Video Reasoning in MLLMs},
  author={Feng, Kaituo and Gong, Kaixiong and Li, Bohao and Guo, Zonghao and Wang, Yibing and Peng, Tianshuo and Wu, Junfei and Zhang, Xiaoying and Wang, Benyou and Yue, Xiangyu},
  booktitle=NIPS,
  year={2025}
}

@inproceedings{clip,
  title={Learning transferable visual models from natural language supervision},
  author={Radford, Alec and Kim, Jong Wook and Hallacy, Chris and Ramesh, Aditya and Goh, Gabriel and Agarwal, Sandhini and Sastry, Girish and Askell, Amanda and Mishkin, Pamela and Clark, Jack and Krueger, Gretchen and Sutskever, Ilya},
  booktitle=ICML,
  year={2021},
}

@inproceedings{aks,
  title     = {Adaptive Keyframe Sampling for Long Video Understanding},
  author    = {Tang, Xi and Qiu, Jihao and Xie, Lingxi and Tian, Yunjie and Jiao, Jianbin and Ye, Qixiang},
  booktitle = CVPR,
  year      = {2025},
}

@inproceedings{qframe,
  title     = {Q-Frame: Query-aware Frame Selection and Multi-Resolution Adaptation for Video-LLMs},
  author    = {Zhang, Shaojie and Yang, Jiahui and Yin, Jianqin and Luo, Zhenbo and Luan, Jian},
  booktitle = ICCV,
  year      = {2025}
}

@article{longvila,
  title={Longvila: Scaling long-context visual language models for long videos},
  author={Chen, Yukang and Xue, Fuzhao and Li, Dacheng and Hu, Qinghao and Zhu, Ligeng and Li, Xiuyu and Fang, Yunhao and Tang, Haotian and Yang, Shang and Liu, Zhijian and He, Yihui and Yin, Hongxu and Molchanov, Pavlo and Kautz, Jan and Fan, Linxi and Zhu, Yuke and Lu, Yao and Han, Song},
  journal={arXiv:2408.10188},
  year={2024}
}

@inproceedings{videoagent,
  title={Videoagent: Long-form video understanding with large language model as agent},
  author={Wang, Xiaohan and Zhang, Yuhui and Zohar, Orr and Yeung-Levy, Serena},
  booktitle=ECCV,
  year={2024},
}

@article{qwen2,
  title={Qwen2 Technical Report},
  author={An Yang and Baosong Yang and Binyuan Hui and Bo Zheng and Bowen Yu and Chang Zhou and Chengpeng Li and Chengyuan Li and Dayiheng Liu and Fei Huang and Guanting Dong and Haoran Wei and Huan Lin and Jialong Tang and Jialin Wang and Jian Yang and Jianhong Tu and Jianwei Zhang and Jianxin Ma and Jin Xu and Jingren Zhou and Jinze Bai and Jinzheng He and Junyang Lin and Kai Dang and Keming Lu and Ke-Yang Chen and Kexin Yang and Mei Li and Min Xue and Na Ni and Pei Zhang and Peng Wang and Ru Peng and Rui Men and Ruize Gao and Runji Lin and Shijie Wang and Shuai Bai and Sinan Tan and Tianhang Zhu and Tianhao Li and Tianyu Liu and Wenbin Ge and Xiaodong Deng and Xiaohuan Zhou and Xingzhang Ren and Xinyu Zhang and Xipin Wei and Xuancheng Ren and Yang Fan and Yang Yao and Yichang Zhang and Yunyang Wan and Yunfei Chu and Zeyu Cui and Zhenru Zhang and Zhi-Wei Fan},
  journal={arXiv:2407.10671},
  year={2024},
}

@article{focus,
  title={Focus: Efficient keyframe selection for long video understanding},
  author={Zhu, Zirui and Xu, Hailun and Luo, Yang and Liu, Yong and Sarkar, Kanchan and Yang, Zhenheng and You, Yang},
  journal={arXiv preprint arXiv:2510.27280},
  year={2025}
}

@inproceedings{blip2,
  title={Blip-2: Bootstrapping language-image pre-training with frozen image encoders and large language models},
  author={Li, Junnan and Li, Dongxu and Savarese, Silvio and Hoi, Steven},
  booktitle=ICML,
  year={2023}
}

@article{mplug,
  title={mplug-owl: Modularization empowers large language models with multimodality},
  author={Ye, Qinghao and Xu, Haiyang and Xu, Guohai and Ye, Jiabo and Yan, Ming and Zhou, Yiyang and Wang, Junyang and Hu, Anwen and Shi, Pengcheng and Shi, Yaya and others},
  journal={arXiv:2304.14178},
  year={2023}
}

@inproceedings{internvl,
  title={Internvl: Scaling up vision foundation models and aligning for generic visual-linguistic tasks},
  author={Chen, Zhe and Wu, Jiannan and Wang, Wenhai and Su, Weijie and Chen, Guo and Xing, Sen and Zhong, Muyan and Zhang, Qinglong and Zhu, Xizhou and Lu, Lewei and others},
  booktitle=CVPR,
  year={2024}
}

@article{instructgpt,
  title={Training language models to follow instructions with human feedback},
  author={Ouyang, Long and Wu, Jeffrey and Jiang, Xu and Almeida, Diogo and Wainwright, Carroll and Mishkin, Pamela and Zhang, Chong and Agarwal, Sandhini and Slama, Katarina and Ray, Alex and others},
  journal=NIPS,
  year={2022}
}

@article{palm,
  title={Palm 2 technical report},
  author={Anil, Rohan and Dai, Andrew M and Firat, Orhan and Johnson, Melvin and Lepikhin, Dmitry and Passos, Alexandre and Shakeri, Siamak and Taropa, Emanuel and Bailey, Paige and Chen, Zhifeng and others},
  journal={arXiv:2305.10403},
  year={2023}
}

@inproceedings{timechat,
  title={Timechat: A time-sensitive multimodal large language model for long video understanding},
  author={Ren, Shuhuai and Yao, Linli and Li, Shicheng and Sun, Xu and Hou, Lu},
  booktitle=CVPR,
  year={2024}
}

@article{minigpt4,
  title={Minigpt4-video: Advancing multimodal llms for video understanding with interleaved visual-textual tokens},
  author={Ataallah, Kirolos and Shen, Xiaoqian and Abdelrahman, Eslam and Sleiman, Essam and Zhu, Deyao and Ding, Jian and Elhoseiny, Mohamed},
  journal={arXiv:2404.03413},
  year={2024}
}

@inproceedings{goldfish,
  title={Goldfish: Vision-language understanding of arbitrarily long videos},
  author={Ataallah, Kirolos and Shen, Xiaoqian and Abdelrahman, Eslam and Sleiman, Essam and Zhuge, Mingchen and Ding, Jian and Zhu, Deyao and Schmidhuber, J{\"u}rgen and Elhoseiny, Mohamed},
  booktitle=ECCV,
  year={2024},
}
\bibliographystyle{iclr2027_conference}

\clearpage
\appendix
\setcounter{tocdepth}{2}
\section*{Appendix}
\addcontentsline{toc}{section}{Appendix}
\startcontents
\printcontents{}{1}{\setcounter{tocdepth}{2}}
\clearpage
\section{Experimental Setup Details}
\subsection{Model Checkpoints}
Table~\ref{tab:ext_checkpoints} lists the models we use and the checkpoints loaded during evaluation. 

\begin{table}[h]
\centering
\small
\setlength{\tabcolsep}{8pt}
\setlength{\belowcaptionskip}{-10pt}
\renewcommand{\arraystretch}{1.12}
\caption{Model checkpoints used in our experiments.}
\begin{tabular}{ll}
\toprule
\textbf{Model name} & \textbf{Checkpoint} \\
\midrule
Qwen2\textendash VL-7B 
& \texttt{Qwen/Qwen2-VL-7B-Instruct} \\

Qwen2.5\textendash VL-3B 
& \texttt{Qwen/Qwen2.5-VL-3B-Instruct} \\

Qwen2.5\textendash VL-7B
& \texttt{Qwen/Qwen2.5-VL-7B-Instruct} \\

Qwen3\textendash VL-4B
& \texttt{Qwen/Qwen3-VL-4B-Instruct} \\

Qwen3\textendash VL-8B 
& \texttt{Qwen/Qwen3-VL-8B-Instruct} \\

Qwen3.5\textendash VL-4B 
& \texttt{Qwen/Qwen3.5-VL-4B} \\

Qwen3.5\textendash VL-9B 
& \texttt{Qwen/Qwen3.5-VL-9B} \\

Qwen3.5\textendash VL-27B 
& \texttt{Qwen/Qwen3.5-VL-27B} \\

LLaVA\textendash Video
& \texttt{lmms-lab/LLaVA-Video-7B-Qwen2} \\

CLIP
& \texttt{ViT-B/32} \\
\bottomrule
\end{tabular}
\label{tab:ext_checkpoints}
\end{table}

\subsection{Datasets}
\label{sec:datasets}
We evaluate our method on two widely used benchmarks for long-video multiple-choice question answering. 

\textbf{VideoMME (w/o sub)}~\cite{videomme} is a human-annotated benchmark containing 900 videos and 2,700 question-answer pairs, with video durations ranging from 11 seconds to 1 hour. It covers 6 domains and 30 fine-grained subcategories, providing a broad evaluation of video understanding ability. Following the \emph{w/o sub} setting, we evaluate using video frames only, without subtitle input. 

\textbf{LongVideoBench (LVB)}~\citep{longvideobench} is a large-scale benchmark for long-context interleaved video-language understanding. It contains 3,763 web-collected videos of up to one hour and 6,678 human-annotated multiple-choice questions, each with four or five candidate options. We use the LongVideoBench validation set downloaded from Hugging Face and rerun all LVB experiments, including baseline comparisons and ablation studies, on this same validation set to ensure consistent evaluation.

\section{Additional Results and Analysis}
\subsection{Performance Breakdown by Video Duration}
\begin{table}[H]
\centering
\small
\caption{\textbf{Accuracy (\%) by video duration on VideoMME} (w/o subtitles) with Qwen2.5-VL-7B. Short, medium, and long videos last less than 2 minutes, 4--15 minutes, and 30--60 minutes, respectively; numbers in parentheses are the number of questions. $\Delta$ denotes the gain of VideoRouter over the stronger single branch. Oracle denotes the accuracy when either branch is correct.}
\label{tab:videomme_duration}
\begin{tabular}{lccc|c}
\toprule
Method & Short (900) & Medium (900) & Long (900) & Overall (2700) \\
\midrule
Local & 68.22 & 58.89 & 53.22 & 60.11 \\
Global & 71.00 & 61.78 & 52.11 & 61.63 \\
Router & \textbf{74.67} & \textbf{65.00} & \textbf{56.89} & \textbf{65.52} \\
$\Delta$ & \textcolor{red}{+3.67} & \textcolor{red}{+3.22} & \textcolor{red}{+3.67} & \textcolor{red}{+3.89} \\
\midrule
Oracle & 78.67 & 72.78 & 63.44 & 71.63 \\
\bottomrule
\end{tabular}
\end{table}
 
Table~\ref{tab:videomme_duration} reports the accuracy on the short, medium, and long subsets of VideoMME, using Qwen2.5-VL-7B as the backbone. As expected, all methods degrade as videos become longer, since a fixed visual budget must cover increasingly more content. More importantly, the relative strength of the two views shifts with video duration. The global branch outperforms the local branch by 2.78 and 2.89 points on short and medium videos, respectively, but falls behind it by 1.11 points on long videos. With a fixed number of representative frames, the temporal coverage of the global view becomes increasingly sparse on long videos, so brief but decisive evidence is more likely to be missed, and query-guided local evidence becomes relatively more valuable. Therefore, the preferred evidence granularity varies not only across benchmarks (Table~\ref{tab:main_result}) but also with video duration within the same benchmark, further indicating that no single fixed view is sufficient. The oracle gap over the stronger branch also grows from 7.67 points on short videos to 11.00 and 10.22 points on medium and long videos, showing that the two views become more complementary as videos get longer. VideoRouter consistently improves over the stronger branch across all three subsets (+3.67, +3.22, and +3.67), and the gain on long videos is as large as that on short videos, demonstrating that the benefit of verification-guided routing does not diminish as video length increases.

\subsection{Performance Breakdown by Task Type}
\begin{table}[h]
\centering
\caption{\textbf{Accuracy (\%) by task type on VideoMME} (w/o subtitles) with Qwen2.5-VL-7B. Numbers in parentheses are the number of questions per task type. $\Delta$ denotes the gain of VideoRouter over the stronger single branch. Oracle denotes the accuracy when either branch is correct.}
\label{tab:videomme_task}
\setlength{\tabcolsep}{3pt}
\resizebox{\textwidth}{!}{%
\begin{tabular}{l*{12}{c}|c}
\toprule
Method & \hd{Action}{Reas.} & \hd{Action}{Rec.} & \hd{Attr.}{Perc.} & \hd{Count-}{ing} & \hd{Info.}{Syn.} & \hd{}{OCR} & \hd{Object}{Reas.} & \hd{Object}{Rec.} & \hd{Spatial}{Perc.} & \hd{Spatial}{Reas.} & \hd{Temp.}{Perc.} & \hd{Temp.}{Reas.} & \hd{}{Overall} \\
 & (285) & (313) & (222) & (268) & (323) & (139) & (454) & (354) & (54) & (56) & (55) & (177) & (2700) \\
\midrule
Local & 52.98 & 59.11 & 71.62 & 38.43 & 75.23 & 67.63 & 59.91 & 64.41 & \textbf{64.81} & 73.21 & 65.45 & 42.94 & 60.11 \\
Global & 54.39 & 62.62 & 71.62 & 40.30 & 78.02 & 69.78 & 58.15 & 65.82 & 62.96 & \textbf{76.79} & 65.45 & 49.15 & 61.63 \\
Router & \textbf{57.19} & \textbf{66.45} & \textbf{74.77} & \textbf{47.39} & \textbf{81.11} & \textbf{76.98} & \textbf{63.44} & \textbf{68.64} & 62.96 & \textbf{76.79} & \textbf{70.91} & \textbf{50.28} & \textbf{65.52} \\
$\Delta$ & \textcolor{red}{+2.81} & \textcolor{red}{+3.83} & \textcolor{red}{+3.15} & \textcolor{red}{+7.09} & \textcolor{red}{+3.10} & \textcolor{red}{+7.19} & \textcolor{red}{+3.52} & \textcolor{red}{+2.82} & \textcolor{gray}{-1.85} & \textcolor{gray}{0.00} & \textcolor{red}{+5.45} & \textcolor{red}{+1.13} & \textcolor{red}{+3.89} \\
\midrule
Oracle & 62.11 & 73.16 & 81.08 & 52.99 & 83.59 & 82.01 & 68.50 & 77.12 & 72.22 & 80.36 & 80.00 & 62.15 & 71.63 \\
\bottomrule
\end{tabular}}
\end{table}
 
Table~\ref{tab:videomme_task} reports the accuracy of the local branch, the global branch, and VideoRouter on the 12 task types of VideoMME, using Qwen2.5-VL-7B as the backbone. The global branch outperforms the local branch on 8 of the 12 task types, and its advantage is most pronounced on questions that require a holistic understanding of the video, such as Temporal Reasoning (49.15 vs.\ 42.94), Action Recognition (62.62 vs.\ 59.11), and Information Synopsis (78.02 vs.\ 75.23). Answering these questions depends on the overall progression of events rather than on a single moment, and therefore benefits from the broad temporal coverage provided by the global view. This observation is consistent with Table~\ref{tab:main_result}, where VideoMME is better served by global coverage. In contrast, the local branch is on par with or better than the global branch on Object Reasoning, Spatial Perception, Attribute Perception, and Temporal Perception, where the answer typically hinges on localized visual details. Overall, VideoRouter improves over the stronger branch on 10 of the 12 task types, with the largest gains on OCR (+7.19) and Counting (+7.09), suggesting that verification-guided routing is particularly helpful when the decisive evidence is sparse and fine-grained.

\subsection{Generalization to Stronger MLLM Backbones}
\label{sec:additional_backbones}
To further evaluate the generalizability of VideoRouter, we extend our experiments to three recent and stronger MLLM backbones, ranging from 8B to 27B parameters. 

\begin{table}[h]
\definecolor{lightblue}{RGB}{235,242,250}
\definecolor{lightgrey}{RGB}{235,235,235}
\centering
\small
\setlength{\tabcolsep}{6.0pt}
\renewcommand{\arraystretch}{1.12}
\caption{Results with additional recent MLLM backbones on VideoMME and LongVideoBench (LVB). VideoRouter consistently improves over both the Local and Global branches across Qwen3-VL-8B, Qwen3.5-VL-9B, and Qwen3.5-VL-27B. Oracle denotes the upper-bound accuracy obtained when either branch gives the correct answer.}
\begin{tabular}{l|l|ccc|c}
\toprule
\multirow{2}{*}{\textbf{Dataset}} & \multirow{2}{*}{\textbf{Model}} & \multicolumn{4}{c}{\textbf{Method}} \\
\cmidrule(lr){3-6}
& & \textbf{Local} & \textbf{Global} & \textbf{Router} & \textbf{Oracle} \\
\toprule
\multirow{3}{*}{\textbf{VideoMME}}
& Qwen3-VL-8B~\citep{qwen3}    & 61.78 & 65.19 & \cellcolor{lightblue}67.52 & 74.52 \\
& Qwen3.5-VL-9B~\citep{qwen35}  & 62.96 & 68.52 & \cellcolor{lightblue} 71.26 & 76.44 \\
& Qwen3.5-VL-27B~\citep{qwen35}   & 67.26 & 72.81 & \cellcolor{lightblue} 75.00 & 79.59 \\
\toprule
\multirow{3}{*}{\textbf{LVB}}
& Qwen3-VL-8B~\citep{qwen3}   & 60.37 & 60.23 & \cellcolor{lightblue}64.48 & 72.05 \\
& Qwen3.5-VL-9B~\citep{qwen35}   & 62.73 & 63.48 & \cellcolor{lightblue}68.55 & 74.63 \\
& Qwen3.5-VL-27B~\citep{qwen35}  & 64.64 & 65.67 & \cellcolor{lightblue}70.63 & 75.71 \\
\bottomrule
\end{tabular}
\label{tab:additional_backbones}
\end{table}

As shown in Table~\ref{tab:additional_backbones}, the Router consistently outperforms both individual branches across all backbones and datasets. 
Compared with the stronger of the Local and Global branches, routing improves accuracy by 2.19--2.74 points on VideoMME and 4.11--5.07 points on LVB. 
These results show that the benefit of coordinating complementary temporal views persists as the underlying MLLM becomes stronger and larger, while the remaining gap to the Oracle indicates further room for improving evidence routing.

\section{Ablation and Routing Analysis}
\subsection{Ablation on the Global Representation Depth}
\begin{table}[H]
\centering
\small
\setlength{\tabcolsep}{7pt}
\renewcommand{\arraystretch}{1.15}
\caption{\textbf{Ablation on the global representation depth \(d_{\mathrm{rep}}\).} We vary the depth used to construct the global representative view and report the resulting performance on VideoMME and LVB. Larger \(d_{\mathrm{rep}}\) leads to a finer global view, which benefits VideoMME but is less effective on LVB.}
\begin{tabular}{lccccc}
\toprule
\multirow{2}{*}{Dataset} & \multirow{2}{*}{$d_{rep}$} & \multicolumn{4}{c}{Methods} \\
\cmidrule(lr){3-6}
&  & Local & Global & Router & Oracle \\
\midrule
\multirow{3}{*}{VideoMME}
& 2  & 60.11 & 61.63 & 65.52 & 71.63 \\
& 3 & 60.11 & 62.96 & 65.59 & 72.00 \\
& 4 & 60.11 & 63.67 & 66.04 & 72.44 \\
\midrule
\multirow{3}{*}{LVB}
& 2  & 60.32 & 57.32 & 64.23 & 69.88 \\
& 3 & 60.32 & 56.66 & 62.23 & 69.13 \\
& 4 & 60.32 & 57.15 & 62.90 & 69.22 \\
\bottomrule
\end{tabular}
\label{tab:drep_ablation}
\end{table}
Table~\ref{tab:drep_ablation} studies the effect of the global representation depth \(d_{\mathrm{rep}}\), which controls the granularity of the representative frames used by the global branch. A larger \(d_{\mathrm{rep}}\) corresponds to a finer global view. On VideoMME, increasing \(d_{\mathrm{rep}}\) consistently improves the global branch, from 61.63 at \(d_{\mathrm{rep}}=2\) to 62.96 at \(d_{\mathrm{rep}}=3\), and further to 63.67 at \(d_{\mathrm{rep}}=4\). The router and oracle results follow the same trend, rising from 65.52 to 65.59 and 66.04, and from 71.63 to 72.00 and 72.44, respectively. This suggests that VideoMME benefits from a finer global representation, likely because a slightly deeper hierarchy preserves broad temporal coverage while providing more informative representative evidence. On LVB, however, the trend is different. The best router performance is achieved at \(d_{\mathrm{rep}}=2\), with an accuracy of 64.23, while increasing the depth to 3 or 4 reduces performance to 62.23 and 62.90, respectively. Although the global branch itself varies only moderately (57.32, 56.66, and 57.15), the final routed result drops more noticeably. This indicates that, for LVB, a coarser global view is more complementary to the local branch, whereas a finer global representation becomes less helpful and may introduce redundancy with the local evidence. Overall, the two datasets exhibit different preferences for global representation depth. VideoMME favors a finer global view, consistent with its stronger reliance on broad storyline understanding, while LVB benefits more from keeping the global branch coarse and leaving fine-grained evidence recovery to the local branch.

\subsection{Ablation on the Local Pruning Threshold $\alpha$.}
\begin{table}[h]
\centering
\small
\setlength{\tabcolsep}{7pt}
\renewcommand{\arraystretch}{1.15}
\caption{\textbf{Effect of the local pruning threshold \(\alpha\).} We vary the pruning threshold of the local branch while keeping the global branch fixed, and report the resulting performance on VideoMME and LVB. A smaller \(\alpha\) yields less strict local pruning. The results show that improving the standalone local branch does not always lead to better routed performance.}
\begin{tabular}{lccccc}
\toprule
\multirow{2}{*}{Dataset} & \multirow{2}{*}{$\alpha$} & \multicolumn{4}{c}{Methods} \\
\cmidrule(lr){3-6}
&  & Global & Local & Router & Oracle \\
\midrule
\multirow{2}{*}{VideoMME}
& 0.95  & 61.62 & 60.11 & 65.52 & 71.63 \\
& 0.90 & 61.62 & 63.11 & 65.59 & 71.41 \\
\midrule
\multirow{2}{*}{LVB}
& 0.95  & 57.32 & 60.32 & 64.23 & 69.88 \\
& 0.90 & 57.32 & 60.40 & 62.23 & 68.72 \\
\bottomrule
\end{tabular}
\label{tab:prune_threshold_ablation}
\end{table}
Table~\ref{tab:prune_threshold_ablation} studies the effect of the local pruning threshold \(\alpha\), which controls how aggressively the local branch filters the leaves of temporal hierarchies. A smaller \(\alpha\) corresponds to less strict pruning and thus retains more local candidates. On VideoMME, relaxing the threshold from \(\alpha=0.95\) to \(\alpha=0.90\) improves the standalone local branch from 60.11 to 63.11, while the routed result changes only marginally from 65.52 to 65.59. At the same time, the oracle slightly decreases from 71.63 to 71.41. This suggests that a less strict local branch can make local reasoning itself stronger, but does not substantially improve the final routed result because its complementarity with the global branch becomes weaker. On LVB, the same relaxation yields only a negligible improvement in the local branch, from 60.32 to 60.40, but causes a clear drop in both router and oracle performance, from 64.23 to 62.23 and from 69.88 to 68.72, respectively. This indicates that, although the local branch is not significantly weakened by stricter pruning, a stricter threshold produces a more complementary local view, which is more beneficial when combined with the global branch. Overall, these results reveal a key trade-off: optimizing the local branch in isolation is not always aligned with optimizing the overall system. A weaker local branch may still be more useful if it contributes evidence that is more distinct from the global branch. This observation further supports our formulation of long-video understanding as coordinating complementary evidence views rather than optimizing a single branch alone.

\subsection{Performance Breakdown by Verification Score Range}
Table~\ref{tab:score_range_breakdown} further analyzes routing behavior under different verifier score pairs. Overall, samples with higher verification scores are usually associated with higher accuracy, while lower-score combinations tend to be less reliable, supporting the role of verification as an indicator of evidence quality.
\begin{table*}[h]
\centering
\small
\setlength{\tabcolsep}{6pt}
\renewcommand{\arraystretch}{1.15}
\caption{\textbf{Performance breakdown by verifier score range on VideoMME and LVB.} Each sample is grouped by the pair of verification scores assigned to the global and local branches. Higher score pairs generally correspond to more reliable branch predictions, while lower-score pairs are more ambiguous and harder to resolve.}
\begin{tabular}{llccccc}
\toprule
\multirow{2}{*}{Dataset} & \multirow{2}{*}{Model} & \multicolumn{5}{c}{Score Range} \\
\cmidrule(lr){3-7}
& & [2,2] & [2,1] & [1,1] & [1,0] & [0,0] \\
\midrule
\multirow{4}{*}{VideoMME}
& Qwen-2.5-VL-3B~\citep{qwen2.5}  & 63.47 & 47.03 & 36.92 & 38.71 & 48.11 \\
& Qwen-2-VL-7B~\citep{qwen2}     & 63.54 & 51.47 & 45.45 & 24.14 & 50.00 \\
& Qwen-2.5-VL-7B~\citep{qwen2.5}   & 72.22 & 52.78 & 60.67 & 52.70 & 47.50 \\
& LLaVA-Video-7B~\citep{llavanextvideo}   & 68.41 & 46.24 & 50.72 & 59.09 & 61.11 \\
\midrule
\multirow{4}{*}{LVB}
& Qwen-2.5-VL-3B~\citep{qwen2.5}  & 62.41 & 51.85 & 50.00 & 51.22 & 43.48 \\
& Qwen-2-VL-7B~\citep{qwen2}   & 62.27 & 68.42 & 51.02  & 44.00 & 43.62 \\
& Qwen-2.5-VL-7B~\citep{qwen2.5}   & 64.43 & 63.27 & 70.91 & 52.87 & 50.33 \\
& LLaVA-Video-7B~\citep{llavanextvideo} & 69.65 & 63.09 & 53.80 & 44.59 & 40.63 \\
\bottomrule
\end{tabular}
\label{tab:score_range_breakdown}
\end{table*}

On VideoMME, the \([2,2]\) group is consistently among the most accurate across all backbones, reaching 72.22 for Qwen-2.5-VL-7B and 68.41 for LLaVA-Video-7B. This suggests that when both branches receive strong support, the selected evidence is usually sufficient for correct answering. By contrast, mixed or low-score groups such as \([2,1]\), \([1,1]\), and \([1,0]\) are generally less stable, indicating that these cases remain more ambiguous even when one branch appears moderately confident. Notably, Qwen-2.5-VL-7B still maintains relatively strong performance on \([1,1]\) and \([1,0]\), suggesting that a stronger QA backbone can better exploit partially supported evidence. On LVB, the same broad trend holds, but the differences across score ranges are more model-dependent. The \([2,2]\) group again performs strongly for all backbones, especially for LLaVA-Video-7B at 69.65. At the same time, some partially supported groups, such as \([2,1]\) for Qwen-2-VL-7B (68.42) and \([1,1]\) for Qwen-2.5-VL-7B (70.91), also achieve high accuracy. This suggests that on LVB, useful evidence is sometimes captured asymmetrically across the two branches, and that even moderate verification scores can still correspond to correct answers when the retrieved local details are highly relevant.

Overall, this analysis supports two conclusions. First, verification scores are informative: strongly supported cases are typically easier and more reliable. Second, the relationship is not strictly monotonic, especially on LVB, where fine-grained evidence may appear in only one branch or only partially satisfy the verifier. This further justifies the need for verification-guided routing rather than relying on a single fixed branch or a hard confidence threshold.

\subsection{Effect of Verifier Choice.}
\begin{table}[h]
\centering
\small
\setlength{\tabcolsep}{4.5pt}
\definecolor{lightblue}{RGB}{235,242,250}
\renewcommand{\arraystretch}{1.12}
\caption{Effect of verifier choice on final routing accuracy. We compare Qwen-2.5-VL-7B and LLaVA-Video-7B as the verifier while keeping the QA backbone fixed. $\Delta$ denotes the improvement obtained by using Qwen-2.5-VL-7B as the verifier.}
\begin{tabular}{lccc}
\toprule
QA Model & Qwen Verifier & LLaVA Verifier & $\Delta$ \\
\midrule
\multicolumn{4}{l}{\textbf{LVB}} \\
Qwen-2.5-VL-7B~\citep{qwen2.5}   & \cellcolor{lightblue}64.23 & 63.64 & +0.59 \\
LLaVA-Video-7B~\citep{llavanextvideo}   & \cellcolor{lightblue}63.81 & 63.56 & +0.25 \\
\midrule
\multicolumn{4}{l}{\textbf{VideoMME}} \\
Qwen-2.5-VL-7B~\citep{qwen2.5}   & \cellcolor{lightblue}65.52 & 64.67 & +0.85 \\
LLaVA-Video-7B~\citep{llavanextvideo}   & \cellcolor{lightblue}64.52 & 64.37 & +0.15 \\
\bottomrule
\end{tabular}
\label{tab:verifier_choice}
\end{table}
Table~\ref{tab:verifier_choice} shows that the choice of verifier consistently affects final routing accuracy. Across both datasets and both QA backbones, using Qwen-2.5-VL-7B as the verifier always outperforms using LLaVA-Video-7B. On LVB, the gain is +0.59 when the QA model is Qwen-2.5-VL-7B and +0.25 when the QA model is LLaVA-Video-7B. On VideoMME, the gains are +0.85 and +0.15, respectively. Although these improvements are modest in magnitude, they are highly consistent. This suggests that verifier quality matters for final performance: even when the QA branches are fixed, a stronger verifier can make more reliable routing decisions. At the same time, the relatively small gains indicate that the verifier mainly acts as a refinement module rather than the dominant source of improvement, while the larger gains under the Qwen QA backbone suggest that better branch predictions can be further amplified by more accurate evidence verification.

\subsection{Analysis of Direct Acceptance vs.\ Fallback.}
\begin{figure}[h]
  \centering
  \includegraphics[width=1.0\linewidth]{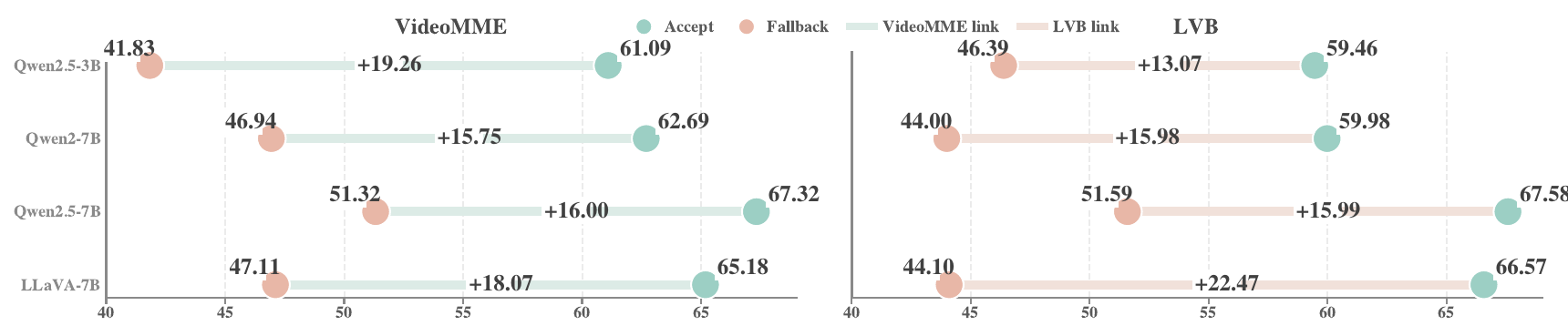}
\caption{\textbf{Accuracy of direct acceptance and fallback reasoning.} 
We report the accuracy of samples that are directly accepted by the verification-guided router (Accept) and those that require an additional union-based reasoning step (Fallback) on VideoMME and LVB. Across all backbones and both benchmarks, directly accepted cases are substantially more accurate than fallback cases, indicating that the verifier successfully identifies predictions supported by stronger evidence.}
\label{fig:acc_vs_fall}
\end{figure}
Figure~\ref{fig:acc_vs_fall} compares the accuracy of directly accepted cases and fallback cases on VideoMME and LVB. A clear pattern emerges across all backbones and both datasets: Accept is consistently much more accurate than Fallback. On VideoMME, the accuracy gap ranges from +15.75 to +19.26 points, specifically +19.26 for Qwen2.5-VL-3B, +15.75 for Qwen2-VL-7B, +16.00 for Qwen2.5-VL-7B , and +18.07 for LLaVA-Video-7B. On LVB, the same trend holds, with gaps of +13.07, +15.98, +15.99, and +22.47 for Qwen2.5-VL-3B, Qwen2-VL-7B, Qwen2.5-VL-7B, and LLaVA-Video-7B, respectively. Notably, the largest gap appears on LLaVA-Video-7B for LVB, where directly accepted samples achieve 66.57 accuracy, while fallback samples drop to 44.10. These results indicate that the verifier is not merely routing samples arbitrarily; rather, it successfully identifies cases in which one branch already provides sufficiently reliable evidence.

\subsection{Fallback Rate.}
\label{sec:fallback_rate}
Since the accuracies of the directly accepted and the fallback subsets are
reported separately in Figure~\ref{fig:acc_vs_fall}, the fraction of
samples that trigger the additional union-based reasoning step is fully
determined by the overall routed accuracy. Let $p$ denote the acceptance rate,
i.e.\ the fraction of samples resolved directly by one branch. The routed
accuracy decomposes as
\begin{equation}
\mathrm{Acc}_{\mathrm{router}}
  \;=\; p \cdot \mathrm{Acc}_{\mathrm{accept}}
  \;+\; (1-p) \cdot \mathrm{Acc}_{\mathrm{fallback}} ,
\label{eq:acc_decomp}
\end{equation}
 
\begin{table}[h]
\centering
\caption{Fallback rate of verification-guided routing.}
\label{tab:fallback_rate}
\begin{tabular}{lcc}
\toprule
Model & VideoMME(\%) & LVB(\%) \\
\midrule
Qwen2.5-VL-3B   & 5.7  & 8.1  \\
Qwen2-VL-7B     & 9.1  & 14.6 \\
Qwen2.5-VL-7B   & 11.3 & 20.9 \\
LLaVA-Video-7B  & 4.5  & 13.4 \\
\bottomrule
\end{tabular}
\end{table}
 
The results are in Table~\ref{tab:fallback_rate}. Two observations follow. First, the fallback rate stays between $4.5\%$ and
$20.9\%$, so between $79\%$ and $95\%$ of all questions are answered by a
single branch whose prediction the verifier accepts directly. The union-based
reasoning pass is thus an exception path rather than the common case. Second,
fallback is systematically more frequent on LVB than on VideoMME for every
backbone, which is consistent with our earlier observation that the global and
local views disagree on more samples on LVB, where fine-grained evidence is
decisive.

\section{Efficiency Analysis}
\subsection{End-to-End Inference Cost}
\begin{table}[h]
\centering
\small
\setlength{\tabcolsep}{10pt}
\renewcommand{\arraystretch}{1.12}
\caption{End-to-end inference cost of VideoRouter with Qwen2.5-VL-7B on VideoMME.}
\begin{tabular}{lc}
\toprule
\textbf{Metric}  & \textbf{VideoMME} \\
\midrule
Accuracy  & 65.52 \\
Avg. visual tokens  & 31{,}889 \\
Latency (sec/question)  & 8.97 \\
Peak GPU memory (GiB)  & 25.27 \\
\bottomrule
\end{tabular}
\label{tab:inference_cost}
\end{table}
Table~\ref{tab:inference_cost} reports the end-to-end inference cost of
VideoRouter with Qwen2.5-VL-7B on VideoMME. VideoRouter averages roughly $30$K visual tokens per question, with a latency of about $9$ seconds and a peak memory
footprint of $25$ GiB. 

\subsection{Cost Comparison with AKS}
\begin{table}[H]
\centering
\caption{Visual token cost of VideoRouter and AKS on LLaVA-Video-7B. AKS uses 5 runs with majority voting, as in Table~\ref{tab:lvb_llava_compare}. \#Frames and \#Tokens are averaged per question.}
\label{tab:token_aks}
\setlength{\tabcolsep}{5pt}
\begin{tabular}{lcccccc}
\toprule
\multirow{2}{*}{Method} & \multicolumn{2}{c}{LVB} & \multicolumn{2}{c}{VideoMME} \\
\cmidrule(lr){2-3} \cmidrule(lr){4-5}
& \#Frames & \#Tokens & \#Frames & \#Tokens \\
\midrule
LLaVA-Video-7B w/ AKS         & 293.5 & 61{,}625 & 318.7 & 66{,}927 \\
LLaVA-Video-7B w/ VideoRouter & \textbf{191.2} & \textbf{40{,}147}  & \textbf{203.7} & \textbf{42{,}784} \\
\bottomrule
\end{tabular}
\end{table}
Table~\ref{tab:token_aks} compares the visual token cost of VideoRouter with AKS on LLaVA-Video-7B under the same MLLM call budget as Table~\ref{tab:lvb_llava_compare}. Following the LLaVA-Video preprocessing, each frame is encoded by SigLIP resulting in $210$ visual tokens per frame. Since AKS aggregates 5 independent VQA runs with $58.69$ and $63.74$ selected frames on average, it consumes $61.6$K and $66.9$K visual tokens per question on LVB and VideoMME, respectively. In contrast, VideoRouter feeds each branch's frames once for VQA and once for verification, and only invokes the additional union-based reasoning pass on samples rejected by the verifier, requiring $40.1$K and $42.8$K visual tokens per question.  hese results show that the gains of VideoRouter do not come from a larger visual budget, but from organizing a smaller set of frames into complementary global and local evidence views and resolving them through verification-guided routing.

\section{Qualitative Analysis}
\subsection{Qualitative Visualization}
\begin{figure}[h]
  \centering
  \includegraphics[width=\linewidth]{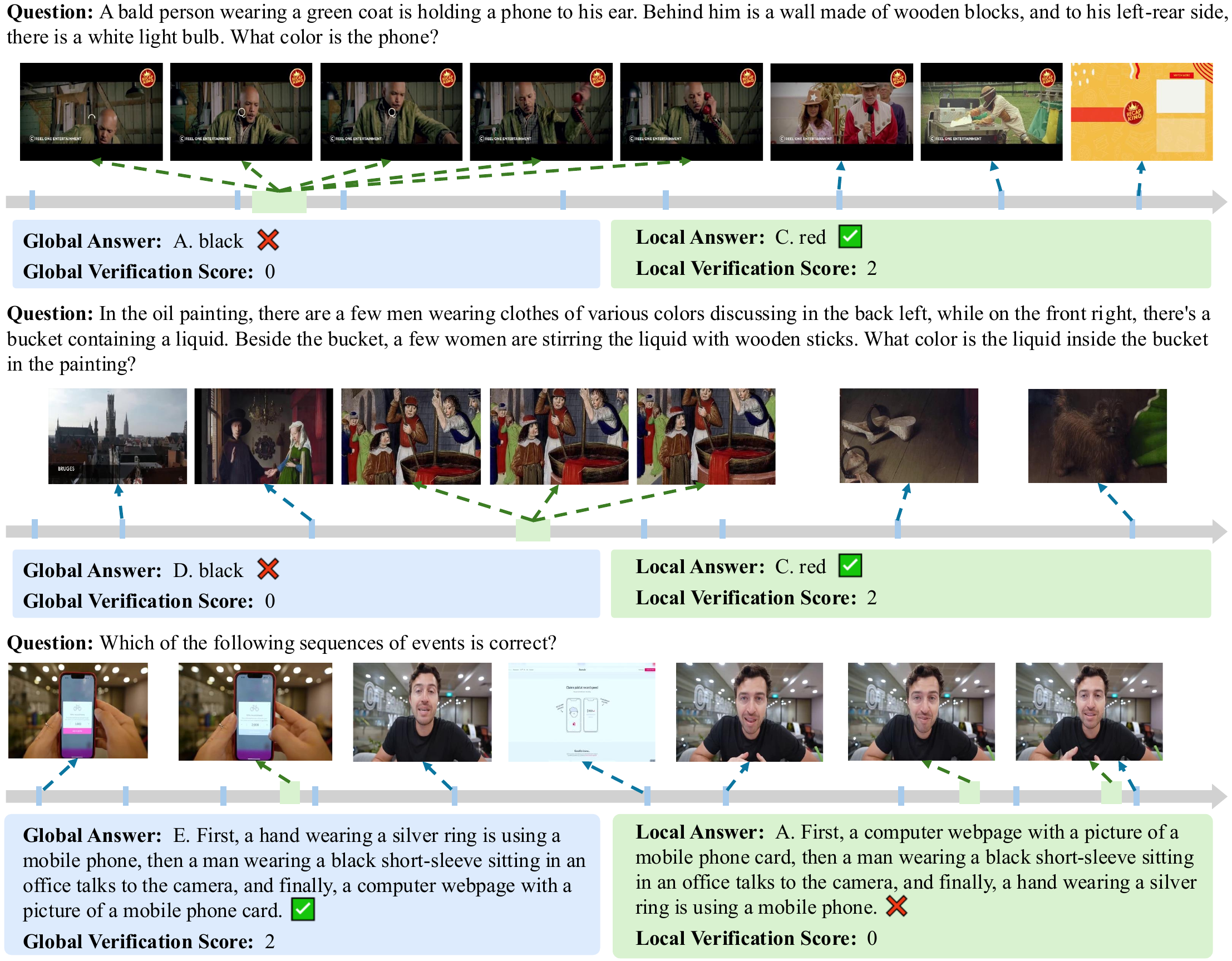}
\caption{\textbf{Qualitative visualization of global--local evidence selection and verification-guided routing.} The gray arrow denotes the temporal axis. Local reasoning is more effective for detail-centric questions, while global reasoning is more reliable for questions involving broader event structure. Verification scores help route the final prediction to the better-supported branch.}
\label{fig:vis}
\end{figure}

Figure~\ref{fig:vis} visualizes the frames selected by the global and local branches along the video timeline, together with their predicted answers and verification scores. It illustrates how the two branches attend to different temporal evidence and how routing prefers the branch whose evidence is better supported. In Case 1, the question depends on a fine-grained visual detail: the phone color. The local branch selects frames where this detail is clearly visible and correctly answers red, whereas the global branch samples broader context but does not capture the evidence needed to identify the color, leading to the wrong answer black. Case 2 shows a similar pattern. The answer hinges on the color of the liquid inside the bucket, and the local branch succeeds because it focuses on frames where the bucket is visible, while the global branch misses these evidence-bearing moments and predicts the wrong color. In Case 3, the question instead requires understanding the temporal order of events. Here the global branch captures the overall event progression and predicts the correct sequence, whereas the local branch attends only to partial moments and fails. Overall, the figure shows that local reasoning is more effective when the answer depends on localized visual evidence, while global reasoning is more reliable when broader temporal structure is required. The router then uses the verification scores to select the branch whose prediction is better supported by the selected frames.



\subsection{Failure Cases}
\begin{table}[h]
\centering
\small
\setlength{\tabcolsep}{4.2pt}
\renewcommand{\arraystretch}{1.1}
\caption{Representative failure cases of verification-guided routing.We group typical routing errors into 3 categories: subtitle/text alignment, temporal order, counting. Global, Local, Router, and GT are shown as option letters for compactness.}
\begin{tabular}{p{1.2cm}|p{8.0cm}|c|c|c|c}
\toprule
\textbf{Failure Reason} & \textbf{Question and Options} & \textbf{Global} & \textbf{Local} & \textbf{Router} & \textbf{GT} \\
\midrule

\textbf{Subtitle}
& \begin{tabular}[h]{@{}p{8.0cm}@{}}
On the yellow floor, a person wearing a blue top is pushing a wooden cart filled with various items. After the subtitles `For the restoration, a lot of it was just cutting acid-free paperboard to the' appear, what does the person in the blue top do? \\
A. Opened the pigment tray \\
B. Wrote with a paintbrush \\
C. Held a red hammer \\
D. Used a brush to apply glue to the items
\end{tabular}
& D & A & D & A \\
\cline{2-6}
& \begin{tabular}[h]{@{}p{8.0cm}@{}}
Before the subtitle says `she quickly hides inside a dead tree with James. The latter wants her to return to' what does the woman in a blue top do? \\
A. She is swimming in the water \\
B. She assists a person in walking \\
C. She is combing her hair \\
D. She wakes up a man lying on the ground\\
E. She is applying lipstick
\end{tabular}
& B & A & A & B \\
\midrule

\textbf{Temporal Order}
& \begin{tabular}[h]{@{}p{8.0cm}@{}}
In a bedroom with two warm-colored table lamps and two paintings hanging on the wall, a short-haired woman wearing an orange outfit is sitting on the bed turning on a computer. What did she do before turning on the computer on the bed?\\
A. Changed into a black coat\\
B. Visited a museum\\
C. Took a shower\\
D. Wrote something in front of the computer\\
E. Put on shoes
\end{tabular}
& A & D & D & C \\
\midrule

\textbf{Counting}
& \begin{tabular}[h]{@{}p{8.0cm}@{}}
How many women appear in the video in total?\\
A. 4 women\\
B. 5 women\\
C. 2 women\\
D. 1 woman\\
E. 3 women
\end{tabular}
&C & C & C & D \\
\bottomrule
\end{tabular}
\label{tab:failure_case_breakdown}
\end{table}
Besides hallucination, Table~\ref{tab:failure_case_breakdown} summarizes three representative failure categories of the proposed router: subtitle, temporal order, and counting. A common pattern across these cases is that neither the global nor the local branch alone captures sufficiently complete evidence, making final routing inherently difficult.

For subtitle questions, the answer depends on precise synchronization between a short-lived visual event and a specific subtitle. In such cases, the global branch often preserves broader context but fails to isolate the exact moment aligned with the target subtitle, while the local branch may focus on visually salient frames that are not temporally aligned with the required subtitle cue. As a result, both branches can appear plausible while still missing the correct text--video correspondence.

For temporal order questions, the difficulty lies in recovering the relative order between multiple events rather than identifying a single relevant moment. The global branch may retain broad event progression but miss decisive local transitions, whereas the local branch may capture one or two relevant clips without covering the full event chain. Consequently, the router may receive partial but incomplete support from both branches, which makes reliable selection challenging.

For counting questions, both branches often suffer from incomplete coverage. The local branch tends to focus on a subset of instances, while the global branch, although broader, may still be too sparse to support accurate counting. In these cases, the failure is not simply that one branch is worse than the other, but that neither branch provides sufficiently exhaustive evidence for an exact numerical answer.

Overall, these failures suggest that routing errors often arise not because one branch is clearly superior, but because the required evidence is temporally aligned, sequentially distributed, or globally accumulative in a way that neither branch fully captures. 

\section{Limitations}
\label{sec:limit}
Despite its effectiveness, our method has several limitations. First, the temporal hierarchy is currently constructed mainly from visual cues. While this is sufficient to provide a useful coarse-to-fine temporal structure, high-level story transitions and event boundaries in long videos are often also signaled by audio and text, such as speech, subtitles, or narration. Incorporating these additional modalities could lead to a more semantically faithful video structure. 
Second, although the proposed verification-guided router consistently improves over using either branch alone, there remains a noticeable gap between the routed results and the oracle upper bound, indicating that branch coordination still can be improved. In particular, the current verifier relies on coarse support levels and may struggle on highly ambiguous cases where evidence is partial, subtle, or distributed across both views. Third, our experiments focus on long-video multiple-choice question answering, and it remains to be seen how well the framework generalizes to more open-ended long-video reasoning settings. We view these limitations as promising directions for future work, including query-adaptive video structuring, stronger evidence verification, and broader evaluation beyond multiple-choice QA.

\clearpage
\section{Additional Method Details}
\subsection{Illustration of Tree Construction}
\label{sec:clustering}
\begin{figure}[h]
  \centering
  \includegraphics[width=\linewidth]{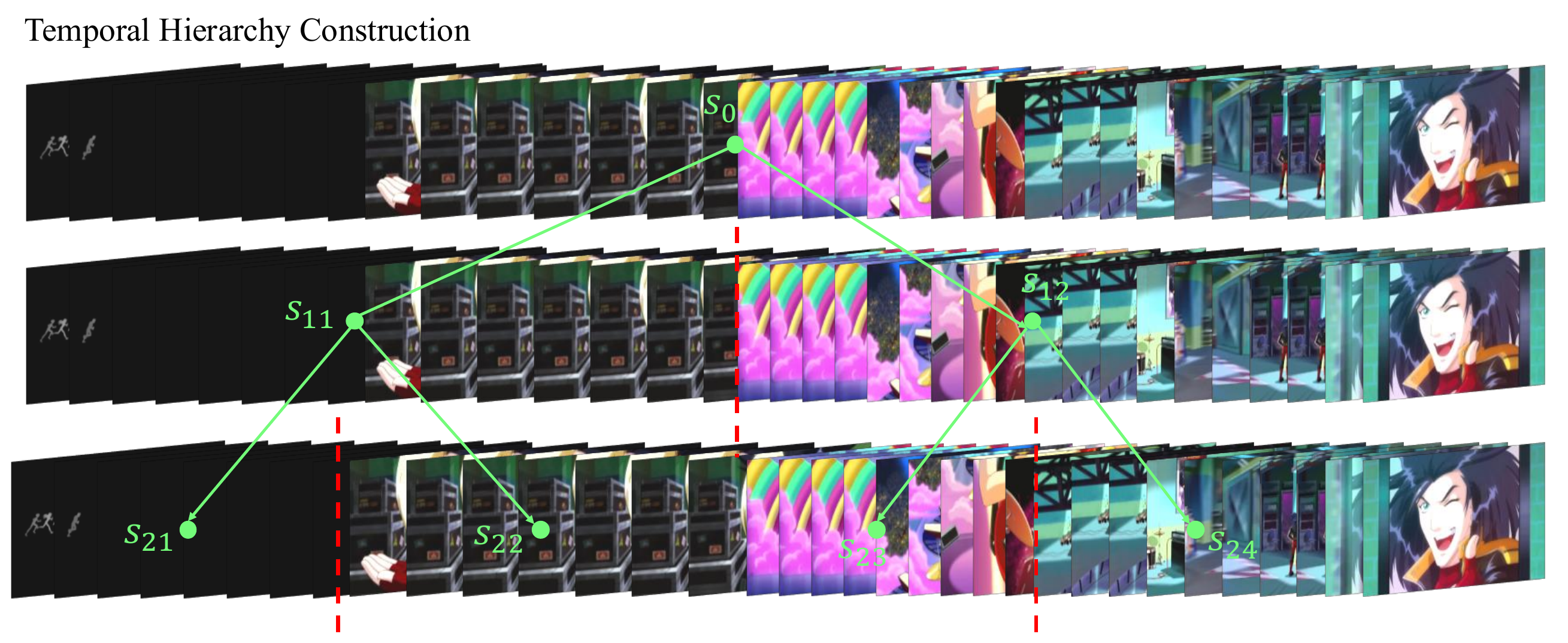}
\caption{Coarse-to-fine construction of the temporal hierarchy. A video is first partitioned into high-level temporally contiguous segments and then recursively refined into finer sub-segments to expose more detailed structure. Red dashed lines indicate temporal boundaries, and green edges denote parent--child relations. Upper levels capture broad storyline structure, while lower levels preserve more localized event details.}
\label{fig:tree}
\end{figure}
Figure~\ref{fig:tree} illustrates the coarse-to-fine construction of the temporal hierarchy. Starting from the full video, we first obtain a high-level temporal partition, where each segments roughly corresponds to a coarse story unit, such as \(S_{11}\). The two subscripts indicate the hierarchy level and the segment index within that level, respectively. For example, in \(S_{11}\), the first subscript \(1\) denotes the first level, and the second subscript \(1\) denotes the first segment at that level.  The decomposition is performed recursively using Algorithm~\ref{alg:seg}, which partitions each non-leaf node into finer temporally contiguous child segments while preserving semantic coherence within each segment. For example, \(S_{11}\) is further decomposed into second-level nodes \(S_{21}\) and \(S_{22}\). As the hierarchy goes deeper, the segments progressively transition from broad story units to more localized event- and clip-level structures, forming a story--event--clip hierarchy. In the figure, red dashed lines indicate the temporal boundaries between sibling segments, while green connections denote the parent--child relations across levels.

\subsection{Prompts}
\label{sec:prompts}
Figure~\ref{fig:qa_prompt} shows the QA prompt format used in our experiments. For LVB questions with five options, the template is extended accordingly (A–E).
\begin{figure*}[h]
\centering
\begin{minipage}{1.00\textwidth}
\begin{promptbox}[frametitle={QA Prompt}]
\ttfamily\small
Select the best answer to the following multiple-choice question based on the video.\\
Respond with only the letter of the correct option.\\[0.5ex]
Question: \{q\}

A. \{o\_1\}\\
B. \{o\_2\}\\
C. \{o\_3\}\\
D. \{o\_4\}\\[0.5ex]
Answer (one of: A, B, C, D):
\end{promptbox}
\end{minipage}
\caption{Prompt template for video multiple-choice QA.}
\label{fig:qa_prompt}
\end{figure*}

\noindent
Figure~\ref{fig:router_prompt} shows the prompt template used by the verifier to assess whether a candidate answer is supported by the sampled video frames.

\begin{figure*}[h]
\centering
\begin{minipage}{1.00\textwidth}
\begin{promptbox}[frametitle={Verifier Prompt}]
\ttfamily\small
You are a verifier whose job is to determine whether a candidate answer is supported by visual evidence.\\
You will be given a question, the answer options, a candidate answer, and a small set of frames sampled from a video.\\
Carefully examine the frames and judge how strongly they support the candidate answer.\\[4pt]

\textbf{Support levels:}\\
- 2 (Strong evidence): The frames provide clear, direct, and sufficient visual evidence for the candidate answer.\\
- 1 (Weak evidence): The frames contain relevant clues, but the evidence is incomplete, ambiguous, or insufficient for a confident conclusion.\\
- 0 (No evidence): The frames do not provide relevant visual evidence to verify the candidate answer.\\[4pt]

\textbf{Guidelines:}\\
- Assign level 2 only when the key evidence is clearly visible and sufficient to justify the answer.\\
- Assign level 1 when partial or indirect clues are present, but certainty is still lacking.\\
- Assign level 0 when no relevant visual clues are visible.\\
- The reason should briefly explain why the assigned level is appropriate based on the observed frames.\\[4pt]

\textbf{Output format:}\\
Return a single-line JSON object with keys \{"level", "reason"\}. Do not output anything else.\\[4pt]

\textbf{Question:} \{question\}\\
\textbf{Options:} \{options\}\\
\textbf{Candidate Answer:} \{candidate\_letter\}\\
\textbf{Return JSON only:}
\end{promptbox}
\end{minipage}
\caption{Prompt template used by the verifier to assess the support level of a candidate answer given sampled video frames.}
\label{fig:router_prompt}
\end{figure*}

\subsection{Pseudo Code for Temporal Clustering}
\label{sec:pseudo_code}
The pseudo-code for our temporal clustering is shown in Algorithm \ref{alg:seg}.
\subsubsection{Overall Purpose of the Algorithm}
The algorithm performs temporal segmentation on a video sequence. Its goal is to divide the sequence into K contiguous, non-overlapping subsegments, each satisfying a minimum-length constraint, in a way that minimizes the total within-segment sum of squared errors (SSE). Once the optimal segmentation is obtained, the algorithm computes the mean feature vector for each subsegment, which is then used as the cluster center for that subsegment.
\begin{algorithm}[h]
\caption{Temporal Clustering for a Single Segment}
\label{alg:seg}
\textbf{Input:}
$X=\{x_1,\dots,x_T\}$,\ $K$,\ $L_{\min}$ \\
\textbf{Output:}
$\mathcal{S}=\{[s_1,e_1],\dots,[s_K,e_K]\}$ \\
\textbf{Center:}
$C=\{\mu_k\}_{k=1}^K$
\begin{algorithmic}[1]
\STATE \textbf{Prefix sums:} $S_x[0]\leftarrow0$, $S_{xx}[0]\leftarrow0$; for $t{=}1..T$:
$S_x[t]\leftarrow S_x[t-1]{+}x_t$, $S_{xx}[t]\leftarrow S_{xx}[t-1]{+}\langle x_t,x_t\rangle$
\STATE \textbf{Interval cost:} 
$\mathrm{SSE}(i,j)=S_{xx}[j]-S_{xx}[i-1]-\frac{\|S_x[j]-S_x[i-1]\|_2^2}{j-i+1}$
\STATE \textbf{Init:} $dp[0][0]\leftarrow0$; otherwise $+\infty$; $prv\leftarrow-1$
\FOR{$k=1..K$}
  \FOR{$t=kL_{\min}..T$}
    \STATE $best\leftarrow+\infty$, $best\_p\leftarrow-1$; $p_{\min}\leftarrow(k{-}1)L_{\min}$, $p_{\max}\leftarrow t{-}L_{\min}$
    \FOR{$p=p_{\min}..p_{\max}$}
      \STATE $cost\leftarrow dp[k{-}1][p]+\mathrm{SSE}(p{+}1,t)$
      \IF{$cost<best$} \STATE $best\leftarrow cost$; $best\_p\leftarrow p$ \ENDIF
    \ENDFOR
    \STATE $dp[k][t]\leftarrow best$; $prv[k][t]\leftarrow best\_p$
  \ENDFOR
\ENDFOR
\STATE \textbf{Backtrack:} $t\leftarrow T$; $\mathcal{S}\leftarrow\emptyset$
\FOR{$k=K..1$}
  \STATE $p\leftarrow prv[k][t]$; prepend $[p{+}1,t]$ to $\mathcal{S}$; $t\leftarrow p$
\ENDFOR
\STATE \textbf{Compute centers:} for $k=1..K$, $\mu_k\leftarrow \frac{1}{e_k-s_k+1}\sum_{t=s_k}^{e_k} x_t$
\STATE \textbf{return} $(\mathcal{S}, C)$
\end{algorithmic}
\end{algorithm}

\subsubsection{Inputs and Outputs}
\paragraph{Inputs} 
The algorithm takes the following inputs:
\begin{itemize}
    \item \textbf{Sequence $X = \{x_1, \dots, x_T\}$}:  
    An ordered collection of frame features.

    \item \textbf{Number of segments $K$}:  
    Specifies the number of contiguous segments into which the sequence will be divided.

    \item \textbf{Minimum segment length $L_{\min}$}:  
    Ensures that each segment contains at least $L_{\min}$ consecutive video frames.
\end{itemize}

\paragraph{Outputs}
The algorithm produces two outputs:
\begin{itemize}
    \item \textbf{Segmentation $\mathcal{S}$}:  
    \[
    \mathcal{S} = \{[s_1, e_1], \dots, [s_K, e_K]\}.
    \]
    Each pair $[s_k, e_k]$ denotes the start index $s_k$ and end index $e_k$ of segment $k$.  
    The segments are contiguous, non-overlapping, satisfy the minimum-length constraint, and together form the optimal partition minimizing the total SSE.

    \item \textbf{Segment centers $C = \{\mu_k\}_{k=1}^K$}:  
    For each segment $[s_k, e_k]$, the center $\mu_k$ is computed as the mean of all frame features within that interval, serving as the cluster center features for that segment.
\end{itemize}
\newpage
\clearpage
\end{document}